\pdfoutput=1
\documentclass[english]{IEEEtran}
\usepackage{babel}
\usepackage[T1]{fontenc}

\usepackage{array}
\usepackage{booktabs}
\usepackage{threeparttable}
\usepackage{multirow}
\usepackage{amsmath}
\usepackage{amssymb}
\usepackage{graphicx}
\usepackage{xcolor}
\usepackage[unicode=true,hidelinks]{hyperref}

\newcommand{\dN}{\mathrm{d}N}
\newcommand{\ours}{DetField}

\begin{document}

\title{Detectable by Design: Learning a Continuous\\
3D Saliency Field from 2D Detector Responses}

\author{Han Sun\IEEEauthorrefmark{1}, Jun Li and Christy Jie Liang%
\thanks{H. Sun, J. Li and C. Liang are with the Faculty of Engineering and
Information Technology, University of Technology Sydney, Ultimo NSW 2007,
Australia. \texttt{\{han.sun@student, jun.li@, jie.liang@\}.uts.edu.au}.}}

\maketitle
\begingroup\renewcommand\thefootnote{\IEEEauthorrefmark{1}}
\footnotetext{Corresponding author.}
\endgroup

\begin{abstract}
The 3D points kept in a map decide whether a camera can be localized against
it, and this choice matters more as the map becomes sparser. Existing 3D keypoint
detectors usually define saliency by semantics, repeatability, reconstruction
or descriptor matchability. None of these definitions asks whether a point will
actually be detected in an image, and as a result most of the points they
select are not detected again in images.

In this paper, we define 3D saliency as \emph{image detectability}: the response of a
2D interest-point detector at a surface point, aggregated over all views in
which the point is visible. We propose \ours{}, a continuous field over 3D space
that takes a point cloud as input and predicts detectability at any location
around it. The supervision is the response of a 2D detector, back-projected
from images onto the point cloud. Multi-scale curvature invariants are passed to the decoder at
every query point, bypassing the voxel pooling of the encoder.

In a localization benchmark that isolates the choice of 3D points, \ours{} outperforms Harris-3D and five learned 3D keypoint methods on objects
and on real indoor scenes. On unseen rooms, it improves localization success by
$7.4$ points over Harris-3D, the strongest baseline. On objects, it also leads
under a 2D detector not used in training. On sparse maps, it stays ahead of Harris-3D on both objects and real
scenes.
\end{abstract}


\section{Introduction}
\label{sec:intro}

\suppressfloats[t]
\begin{figure}[t]
\centering
\includegraphics[width=\columnwidth]{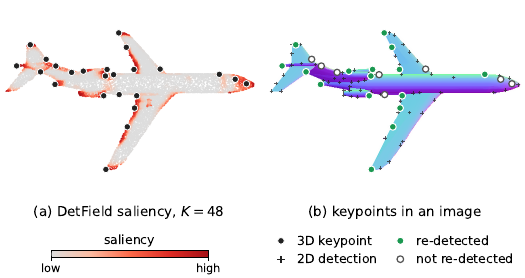}
\caption{\label{fig:teaser}Image detectability on a held-out airplane. (a) The
saliency predicted by \ours{} on the point cloud, shown as the rank within the
object, and the $K{=}48$ keypoints selected from it. (b) The same keypoints
projected into a rendered view, together with the SuperPoint detections in this
view. Most keypoints are re-detected by the 2D detector, which is the property
that \ours{} is trained to predict.}
\end{figure}

A common approach to camera localization against a 3D map is to match 2D
detections in the query image to 3D points of the map and then solve for the
pose. The 3D points kept in the map therefore decide whether localization
succeeds. This is especially true for sparse maps, where each view contains only
a few map points, and every point that is not detected again in the query image
removes a possible correspondence for pose estimation. A 3D keypoint detector for
localization should therefore select the points that a 2D image detector will
detect again in the query image.

However, existing 3D keypoint detectors are not designed for this. They define
saliency by what human annotators consider meaningful~\cite{you2020keypointnet},
by what makes points repeatable under rigid motion~\cite{li2019usip} or
sufficient to reconstruct the shape~\cite{you2022ukpgan}, by what a 3D
descriptor can match between point clouds~\cite{bai2020d3feat}, or by what a
shared image--point embedding can align~\cite{feng20192d3d,wang2021p2}. None of
these definitions considers how the surface will appear to a 2D image detector.
In our experiments on both objects and real scenes, the points selected by
learned 3D keypoint detectors are re-detected in images at about the same rate
as uniformly random surface points. On objects, for example, $11$--$15\%$ of
their keypoints are re-detected, compared with $13\%$ for random points
(Section~\ref{sec:exp}).

In this work, we define 3D saliency directly as \emph{image detectability}: the
response of a 2D image detector at a surface point, aggregated over the views
from which the point can be seen (Fig.~\ref{fig:teaser}). We learn this quantity as a continuous field
over 3D space, named \ours{}, which takes only a point cloud as input. Discrete
3D keypoints are then obtained as the local maxima of this field. Since no image
is needed at test time, \ours{} can be applied to maps built from geometry
alone, such as laser scans. During training, the supervision comes from running
a 2D detector on images of the training data. We use
SuperPoint~\cite{detone2018superpoint} as the 2D detector, and the construction
itself is not specific to SuperPoint. On objects, these images are
renderings of the geometry from many views. On real scenes, they are real
photographs, so the supervision follows the definition exactly and does not rely
on rendering as an approximation.

The design of the model follows two observations. First, among classical
operators, Harris-3D comes closest to predicting where a 2D detector fires. Its
response is a fixed function of two curvature invariants of the surface, and one
of these invariants is also the limit of a multi-view aggregate of normal-map
structure tensors. We therefore feed both invariants at several scales and let
the network learn the combination that a 2D detector actually responds to.
Second, on real scenes a voxel of the encoder is coarser than the neighbourhood
over which these invariants vary. For this reason, we pass them to the decoder
at every query point, bypassing the voxel pooling.

A field trained on the responses of one 2D detector may simply memorize that
detector. We test this on objects by evaluating with D2-Net, a 2D detector and
descriptor never used in training. Under D2-Net, \ours{} keeps its lead and also
scores above the back-projected SuperPoint responses that it is trained on.
Such a result is unlikely if \ours{} only memorized where SuperPoint fires. It
indicates that \ours{} has learned which surface structures a 2D detector
responds to.

Our contributions are summarized as follows:
\begin{itemize}
\item a definition of 3D saliency as image detectability, predicted by a
continuous field that takes only a point cloud as input and is supervised by 2D
detector responses on rendered images for objects and on real photographs for
scenes (Section~\ref{sec:method});
\item per-query curvature invariants derived from the structure of Harris-3D,
which are passed to the decoder around the voxel encoder and give the field
access to the local differential geometry that a 2D detector responds to
(Section~\ref{subsec:qfeat});
\item a localization benchmark in which all methods share the same pipeline and
differ only in the 3D keypoints they select from the map. On this benchmark,
\ours{} outperforms Harris-3D and five learned 3D keypoint methods on objects and
on real indoor scenes queried with real photographs. On unseen rooms, the margin
in localization success is $7.4$ points over Harris-3D and $37$--$46$ points over
the learned methods. On objects, \ours{} also keeps its advantage under a 2D
detector it has never seen (Section~\ref{sec:exp});
\item an evaluation on sparse maps, where \ours{} stays ahead of Harris-3D on
objects at every spacing and on real scenes at spacings of $5$ and $10$\,cm
(Section~\ref{subsec:exp_sparse}).
\end{itemize}

\section{Related Work}
\label{sec:related}

\subsection{Saliency defined by human annotation}

Existing 3D keypoint methods differ mainly in how they define saliency. The most
direct definition comes from human annotation.
KeypointNet~\cite{you2020keypointnet} collects human-annotated keypoints on
ShapeNet objects, and detectors trained fully or partly on these
labels~\cite{qi2017pointnet,qi2017pointnet++,wang2019dynamic,jin2024keypointdetr,back2024to3d}
predict points that people agree are meaningful, such as the tip of a wing or
the corner of a seat. Such points are not necessarily
geometrically distinctive, and a 2D image detector may not detect them in an
image.

\subsection{Saliency defined by the 3D data alone}

Self-supervised 3D keypoint
methods~\cite{li2019usip,you2022ukpgan,zohaib2023sc3k,hou2024keygrid} define
saliency without human labels. They select keypoints through objectives on the
3D data itself, such as repeatability under rigid motion or reconstruction of
the shape. Classical operators such as
ISS~\cite{zhong2009iss}, Harris-3D~\cite{sipiran2011harris} and
NARF~\cite{steder2010narf} also score points from the 3D data alone, and
Harris-3D remains a strong baseline in our experiments. None of these
definitions considers how the surface will appear in an image.

\subsection{Saliency defined by descriptor matchability}

Instead of objectives on a single point cloud, detector-and-descriptor methods
for point cloud
registration~\cite{yew20183dfeat,bai2020d3feat,huang2021predator} define
saliency through matching between two point clouds. They learn a detector
jointly with a 3D descriptor, so that the selected points can be described
distinctively and matched across scans. Among the definitions discussed so far,
this one is the closest to ours, since saliency also depends on whether a point
can be matched. However, their saliency is decided by a 3D descriptor,
whereas the saliency of \ours{} is decided by a 2D image detector.

\subsection{Saliency defined by image-to-point matching}

Cross-modal methods match an image to a point cloud instead of matching two
point clouds. They learn a shared embedding so that image patches and 3D points
can be matched directly~\cite{feng20192d3d,wang2021p2}. P2-Net~\cite{wang2021p2}
also detects keypoints in both modalities, so its saliency is decided by what
the shared embedding can align. This definition involves images, but it depends
on an embedding that P2-Net learns itself. In contrast, the saliency of \ours{}
is decided by an off-the-shelf 2D image detector, which is fixed and independent
of \ours{}. These methods also address a different step of
localization. They match image patches to 3D points, while \ours{} selects the
3D points to be matched. The keypoints selected by \ours{} are therefore
complementary to these matching methods.

\subsection{Lifting 2D responses onto 3D surfaces}

None of the definitions above is based on the response of a 2D image detector.
\ours{} obtains such a definition by lifting 2D detector responses onto the 3D
surface. A common strategy for this is to render the shape from many
viewpoints, apply a 2D operator to each rendering and aggregate the results back
onto the surface. Applications of this strategy include mesh
saliency~\cite{song2021mesh}, shape segmentation~\cite{kalogerakis2017projective},
feature distillation from 2D foundation models~\cite{dutt2024diff3f}, and
few-shot 3D keypoint detection with back-projected semantic
features~\cite{back2024to3d}.
Among these methods, B2-3D~\cite{back2024to3d} is the closest prior work in
how the supervision is constructed. It
back-projects semantic features, while \ours{} back-projects the response of a
2D interest-point detector. \ours{} therefore predicts whether a 2D detector
will respond to a region.

In these works, the geometric meaning of the aggregated response is usually left
implicit. When the 2D response is the trace of the structure tensor of a
rendered normal map, the aggregate has a closed-form limit. We use this limit in
Section~\ref{subsec:geometry} to interpret what \ours{} learns.

\subsection{Continuous saliency fields}

Most of the methods above output a discrete set of keypoints. In contrast,
\ours{} is an implicit field that predicts saliency at any 3D location, and its
architecture follows the convolutional occupancy
network~\cite{peng2020convolutional}. SNAKE~\cite{li2022snake} also predicts
saliency at arbitrary 3D locations from a convolutional point-cloud encoder.
However, SNAKE defines saliency through a combination of self-supervised
objectives, and the resulting saliency has no external reference against which
it can be verified. The saliency of \ours{} is defined by the response of a 2D
detector that is independent of \ours{}. This definition can therefore be
tested with a different 2D detector (Section~\ref{subsec:exp_d2net}).

\section{Method}
\label{sec:method}

\subsection{A continuous field of image detectability}
\label{subsec:field}

Let $X=\{x_i\}_{i=1}^{N}\subset\mathbb R^3$ be an observed point cloud. We learn
a function
\begin{equation}
f_\theta:\mathbb R^3\times\mathcal X\rightarrow[0,1],
\qquad
(q,X)\mapsto \hat s(q\,|\,X),
\end{equation}
which assigns a scalar $\hat s$ to any query location $q$. This scalar is the
image detectability defined in Section~\ref{sec:intro}. It is the response that
a 2D interest-point detector is expected to produce at the surface point nearest
to $q$, aggregated over the views from which that point is visible. We call the learned function \ours{}. It is
defined at any location in 3D space, including locations between the input
points, and discrete keypoints are recovered from it as local maxima
(Section~\ref{subsec:extract}).

\begin{figure*}[t]
\centering
\includegraphics{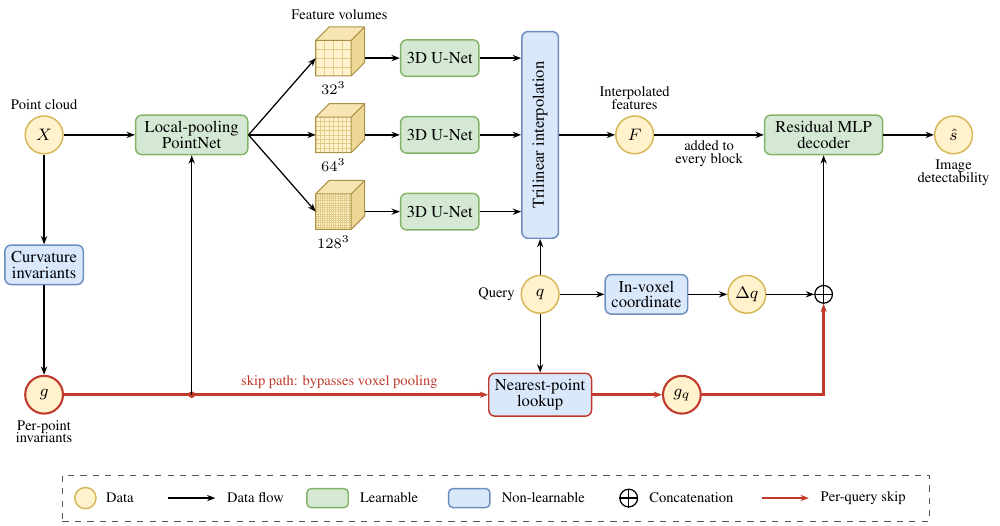}
\caption{\label{fig:model}Architecture of \ours{}. A local-pooling PointNet
encodes the point cloud $X$, together with the per-point curvature invariants $g$
of Section~\ref{subsec:qfeat}, into feature volumes at three resolutions. Each
volume is refined by a 3D U-Net. At a query $q$, the features of the three
volumes are trilinearly interpolated and added to every block of a residual MLP
decoder. The input of the decoder is the in-voxel coordinate $\Delta q$
concatenated with the invariants $g_q$ of the nearest input point. This path
(red) bypasses voxel pooling. The decoder outputs the image detectability
$\hat s$ at $q$.}
\end{figure*}

The architecture follows the convolutional occupancy network
design~\cite{peng2020convolutional}. A local-pooling PointNet encoder maps $X$
to feature volumes at three resolutions, $32^3$, $64^3$ and $128^3$. Each volume is refined
by a three-level 3D U-Net~\cite{cciccek20163dunet}. The three resolutions provide both coarse context and
fine detail without the memory cost of a single wide fine grid. The input of the
residual MLP decoder is the coordinate of $q$ \emph{relative to its voxel},
concatenated with the per-query geometric features of
Section~\ref{subsec:qfeat}. Each decoder block adds the trilinearly interpolated
features of all three volumes at $q$, and the last block predicts a logit
(Fig.~\ref{fig:model}). We use the in-voxel coordinate because we observed that the
decoder memorizes positions from absolute coordinates when the training set is
small.
The remaining hyperparameters, including the input normalization and the number
of encoder points, are listed in the supplementary material (Sec.~S5-A).

\subsection{Per-query geometric features}
\label{subsec:qfeat}

\ours{} is trained to predict where a 2D interest-point detector responds, and
a classical answer to this question is a local differential quantity.
Harris-3D~\cite{sipiran2011harris} is the strongest baseline in our experiments.
It fits a quadric $z=f(x,y)$ to the $k$ nearest neighbours of a point in a local
frame whose $z$-axis is the estimated normal, and scores the point by
$\det E-0.04\,(\operatorname{tr}E)^2$, with $E=\nabla f\nabla f^{\top}+2H^2$ and
$H$ the Hessian of the fit. In this frame the gradient at the point is
approximately zero, so $E\approx2H^2$ and
\begin{equation}
R_{\mathrm{H3D}} \approx 4\bigl[(\kappa_1\kappa_2)^2 - 0.04\,(\kappa_1^2+\kappa_2^2)^2\bigr].
\label{eq:h3d}
\end{equation}
This response is a fixed function of two curvature invariants. The first
invariant is $\|H\|_F^2=\kappa_1^2+\kappa_2^2$, which equals $\|\dN\|_F^2$. Up to
a constant factor, it is also the limit of the multi-view aggregate of
normal-map structure tensors (Section~\ref{subsec:geometry}). The second
invariant is $(\det H)^2=(\kappa_1\kappa_2)^2$, which separates a corner from an
edge.

Instead of this fixed function, \ours{} takes the invariants themselves as input
and learns how to combine them. For every input point, we compute $\|H\|_F^2$,
$(\det H)^2$ and the surface variation $\lambda_{\min}/\sum\lambda$ of the
neighbourhood covariance, each at $k=16$, $32$ and $64$ neighbours. Each of the
nine channels is rank-normalized to $(0,1]$ within the shape, which makes the
features invariant to scale and to sampling density. The features enter the
encoder together with the coordinates. They also enter the decoder at every
query, together with the in-voxel coordinate. A query that does not coincide
with an input point takes the features of its nearest input point, which is the
same rule that assigns its training target (Section~\ref{subsec:loss}).

The path to the decoder is needed because the encoder cannot carry these
quantities to the decoder by itself. On the median scene crop, a voxel of the
finest grid spans about 3.5\,cm. The $32$ nearest points used for the invariants
lie within about 1.8\,cm of the query point. The variation of the invariants
inside a voxel is therefore averaged away by the voxel pooling.
Section~\ref{subsec:exp_ablation} measures the contribution of each path.
The features are computed in the same way for objects and for scenes.

\subsection{Supervision on objects}
\label{subsec:supervision}

For objects, the supervision is built offline for each shape. It uses neither
texture nor human annotation. We use SuperPoint~\cite{detone2018superpoint} as the
2D detector for objects and for real scenes
(Section~\ref{subsec:supervision_scene}). The construction below is not
specific to SuperPoint.

\begin{figure}[t]
\centering
\includegraphics[width=\columnwidth]{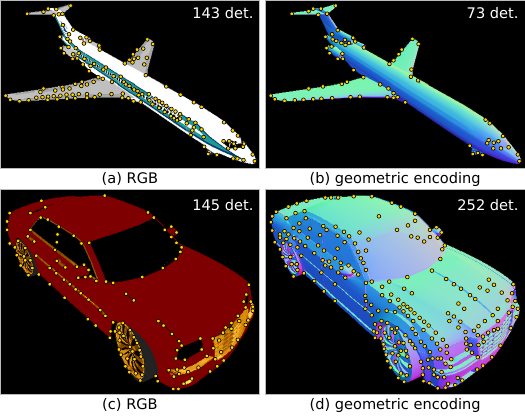}
\caption{\label{fig:encoding}SuperPoint detections on unlit RGB renderings
(a,~c) and on the geometry-only encoding of~\eqref{eq:encoding} (b,~d), from the
same view in each row. The numbers are detection counts. SuperPoint runs on the
grayscale encoding, and (b,~d) show the same views as colour-coded normals for
display. On the airplane, whose flat
surfaces carry a painted livery, the RGB detections follow the paint (window
band, stripes), while the encoding responds to structure. On the uniformly
painted car body, RGB responds mainly at the silhouette and colour boundaries,
whereas the encoding also responds to the geometric detail inside it.}
\end{figure}

\paragraph{Geometry-only image encoding}
With texture, a 2D detector also responds to painted patterns that do not
correspond to any change in shape. We therefore render two geometric quantities
and combine them into a single-channel image,
\begin{equation}
I(u) = 0.7\,n_z(u) + 0.3\,d_{01}(u),
\label{eq:encoding}
\end{equation}
where $n_z=\sqrt{1-n_x^2-n_y^2}$ is computed from the unit surface normal $n$
in camera coordinates, and $d_{01}$ is the
rendered depth affinely mapped to $[0,1]$ using the bounding sphere of the
shape. The image is replicated to the three input channels of the 2D detector.
The normal enters only through $n_z$, so the encoding is invariant to rotation
about the optical axis, and the response does not change when the camera rolls.
Fig.~\ref{fig:encoding} compares detections on RGB renderings and on this
encoding. On textured surfaces, RGB renderings produce responses in flat painted
regions, while the geometric encoding keeps only the structure.

\paragraph{View configuration}
We render each shape from $L=128$ views on a Fibonacci lattice over the sphere.
The camera distance of each view is chosen so that the projected object fills a
fixed fraction of the image. Fixing the apparent size removes object scale from
the definition of saliency. Without it, a view that sees the object end-on would
measure the response at a different spatial scale than a broadside view. The
view layout is illustrated in the supplementary material (Sec.~S5-B).

\paragraph{Per-view response}
We run the 2D detector on each encoded view and keep
its dense detection map. This map is the per-pixel probability before any threshold
or non-maximum suppression. Two corrections are applied before aggregation.
First, a learned 2D detector is in general not equivariant to mirroring. A left--right flip of a
geometric image is a valid view of the mirrored shape, so the difference between
the two responses is noise. We remove this difference by
averaging the response with the mirrored response of the mirrored image, which
gives the symmetrized response $r$. Second, the 2D detector also responds at depth
discontinuities. There the response comes from the silhouette of the surface
behind, so it does not describe the surface at the pixel. We attenuate responses
near occlusion boundaries and strengthen the others, using a local occlusion
measure that separates true boundaries from slanted but continuous surfaces,
such as a wing seen edge-on. This gives the reweighted response $r'$. The exact
form of both corrections is given in the supplementary material (Sec.~S5-B).

\begin{figure*}[t]
\centering
\includegraphics[width=\textwidth]{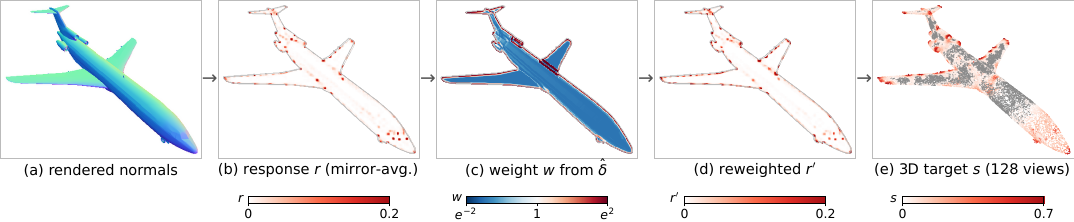}
\caption{\label{fig:supervision}Construction of the object supervision for one
view. (a)~Rendered view as colour-coded normals (detector input:
$I$ of~\eqref{eq:encoding}). (b)~Dense SuperPoint response $r$, averaged with
the mirrored response of the mirrored image. (c)~Reweighting factor derived from
the local occlusion measure (log scale). (d)~Reweighted response $r'$.
(e)~The target $s$ on the point cloud after back-projection and averaging over
all $128$ views. Points with $s$ below $4\%$ of the colour range are shown in
grey. Response maps in (b,~d) are max-pooled over $3\times3$ pixels for
display.}
\end{figure*}

\paragraph{Back-projection and aggregation}
Each surface point is projected into every view, and it is visible in the view
when its depth agrees with the rendered depth to within $0.02$ in normalized
coordinates. The response is read as the maximum over a $9\times9$ pixel
neighbourhood of the projection. Reading a single
pixel discards most of the signal, because the response peak drifts by a few
pixels between views.
Observations with a large local depth spread are also discarded
(supplementary material, Sec.~S5-B). The target scalar is the mean of
the remaining observations,
\begin{equation}
s(x) = \frac{1}{|\mathcal V(x)|}\sum_{l\in\mathcal V(x)} r'_l(x),
\label{eq:aggregate}
\end{equation}
with $\mathcal V(x)$ the set of retained views. We also keep the number $n(x)$
of views in which $x$ is visible before the depth-spread filter. This number is
used as a reliability weight during training.

\subsection{Supervision on real scenes}
\label{subsec:supervision_scene}

On real scenes, the 2D detector can be run on real photographs of the scene. The
supervision then follows the definition of image detectability directly and does
not depend on rendering.

\paragraph{Scene crops}
We use SceneFun3D~\cite{delitzas2024scenefun3d}, whose rooms are captured both by
a Faro laser scanner and by a hand-held camera registered to the scan to within
a few millimetres. Each room is reconstructed once as a Poisson mesh, with normals
oriented towards the cameras that observed them. A crop is the part of this mesh
that is co-visible from one cluster of nearby camera frames. Each crop is then
sampled into a point cloud. The clustering and selection rules are given in the
supplementary material (Sec.~S5-C).

\paragraph{Rendered target}
The object pipeline of Section~\ref{subsec:supervision} is applied to each crop
mesh. Cropping and scan holes leave open mesh boundaries, and the 2D detector
would respond to their silhouettes. Responses near an open boundary are therefore
attenuated. The distance to the boundary is measured along the surface, so that
a hole behind an object does not suppress the object in front of it.

\paragraph{Photographic target}
For each training crop, we select 60 real frames from the video of the same room
by how much of the crop they see. SuperPoint is run on these photographs at half
resolution. A point is visible in a frame when its depth agrees with the depth
map of the frame to within $5$\,cm. For every visible point, the response is read
as the maximum over a $9\times9$ window around its projection, as
in~\eqref{eq:aggregate}. The photographic target $s_{\mathrm{ph}}(x)$ is the mean
over the frames that observe $x$. Photographs are taken from the training rooms
only, and no image of a test room is used for supervision. In a typical crop,
75\% of the points are observed in at least one photograph. The remaining points
keep the rendered target.

\paragraph{Training target}
The photographic and the rendered targets have different scales, so we combine
them by rank. Let $\rho[\cdot]$ be the rank within a set, normalized to $(0,1]$.
The observed points are ordered by $\rho[s_{\mathrm{ph}}]$, and the unobserved
points are ordered by $\rho$ of the rendered target. Each rank is computed on its
own subset. The two subsets are then merged by these normalized ranks, so that a
point at a given percentile of the observed points is treated as being as
salient as a point at the same percentile of the unobserved points. The merged ranks are then mapped monotonically onto the value
distribution of the rendered target. In this way, the ordering comes from the
photographs wherever they exist, and the values of the target have the same
scale as on objects. The loss of Section~\ref{subsec:loss} therefore behaves in
the same way for objects and for scenes. Section~\ref{subsec:exp_ablation}
compares this target with the rendered target alone.

\begin{figure}[t]
\centering
\includegraphics[width=\columnwidth]{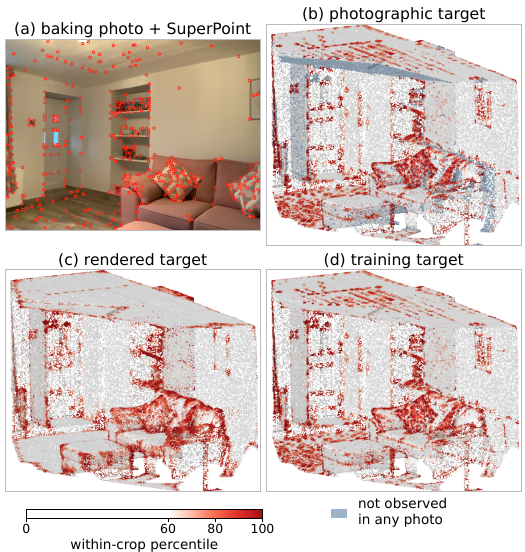}
\caption{\label{fig:scene_target}The real-scene training target on a training
crop. (a)~One of the $60$ photographs used for baking, with its SuperPoint
detections. (b)~Photographic target $s_{\mathrm{ph}}$: mean SuperPoint response over the
photographs that observe each point. Blue-grey points are observed by none. (c)~Rendered target from the encoded renderings of the crop mesh.
(d)~Training target: observed points ordered by (b), the others by (c), mapped
onto the value distribution of (c). Colours show within-crop percentiles.}
\end{figure}

\subsection{Geometric interpretation of the supervision}
\label{subsec:geometry}

The construction of Section~\ref{subsec:supervision} follows a general
procedure. A shape is rendered from many views, a 2D operator is applied to each
view, and the responses are aggregated on the surface. The limit of this
procedure can be computed in closed form when the per-view response is the trace
of the structure tensor of the rendered normal map. Under orthographic
projection and uniformly distributed views, this limit is
$\frac23\|\dN_q\|_F^2=\frac23(\kappa_1^2+\kappa_2^2)$, which is $\frac43$ times
the squared curvedness~\cite{koenderink1992surface}. The same spherical moments also give the variance of an
estimate from a finite number of views, and a correction for view sets that are
non-uniform or partly occluded. Section~S1 of the supplementary material states
these results and sketches their proofs.

A learned 2D detector falls outside this analysis. When SuperPoint responses are
aggregated over the same views, the aggregate correlates positively with
curvedness, but clearly less than the trace response does. The rank correlations
range from $0.34$ to $0.87$, depending on the surface. The field learned by
\ours{} therefore follows curvature only \emph{partially}. It inherits the
ordering corner $>$ edge $>$ flat, and departs from this ordering where the
preferences of the 2D detector depart from geometry. In particular,
$\|\dN\|_F^2$ alone identifies the most detectable points less well than
Harris-3D, which also contains the corner term $(\kappa_1\kappa_2)^2$. This is
why the input features of Section~\ref{subsec:qfeat} include the corner term in
addition to $\|\dN\|_F^2$.

\subsection{Training objective}
\label{subsec:loss}

Queries are sampled as a mixture of near-surface and global locations. For a
query $q$, let $x_q$ be its nearest input point and $d_q$ the distance between
them. The target of $q$ is a shell around the surface,
\begin{equation}
t(q) = s(x_q)\,\exp\!\Bigl(-\frac{d_q^2}{2\sigma^2}\Bigr),
\qquad \sigma = 0.01.
\label{eq:shell}
\end{equation}
Both $d_q$ and $\sigma$ are measured in normalized coordinates. Queries with
$d_q>2\sigma$ are excluded from the loss. Each query is weighted by
\begin{equation}
w(q) = \min\bigl(n(x_q)/c_0,\;1\bigr),\qquad c_0=6.
\label{eq:weight}
\end{equation}
Points seen from few views have an aggregate~\eqref{eq:aggregate} estimated from
little evidence, and this weight lets them contribute less. The loss is the
weighted binary cross-entropy
\begin{equation}
\mathcal L=\frac{\sum_q w(q)\,\mathrm{BCE}\bigl(\hat s(q),t(q)\bigr)}{\sum_q w(q)}.
\label{eq:loss}
\end{equation}

We do not add a peakiness or sparsity regularizer. The
aggregate~\eqref{eq:aggregate} is already concentrated in compact regions where
the 2D detector responds, so an additional local-contrast term is not needed.

\paragraph{Schedule and augmentation}
The object model is trained in two stages. The first stage uses the coordinates
alone, and the second stage adds the per-query features. The scene model is
initialized from the first-stage object model. It is trained with full $SO(3)$
rotations, small translations, coordinate noise and random subsampling of the
input points. Scenes have fewer training samples than objects, and without this
augmentation the scene model overfits within a few dozen epochs. The schedules
and constants are listed in the supplementary material (Sec.~S5-D).

\subsection{Extracting keypoints}
\label{subsec:extract}

For a budget of $K$ keypoints, we evaluate the field at the input points and sort
them by response. Greedy non-maximum suppression with a 3D radius then keeps the
highest-scoring point, removes all points within the radius, and repeats until
$K$ points are kept. The radius is $0.02$ in normalized coordinates.

The scene model is trained with random rotations, so on scene crops we also
average over rotations at test time. The within-crop rank of the field is
averaged over the input and three random rotations of it, each with two random
encoder subsamples. This averaging uses no information beyond the point cloud.

\subsection{Sparse maps}
\label{subsec:sparse}

Maps used for localization are often stored at a fixed voxel resolution of a few
centimetres, which is far coarser than the scans they come from. At this density,
the input points no longer lie where a 2D detector would respond. A method that
can only rank the stored points is then limited by where these points happen to
be. A continuous field can be evaluated anywhere, so we also test \ours{} on maps
thinned by voxel subsampling. The subsampling keeps one point per voxel of edge
$s$, and the per-query curvature invariants are recomputed on the sparse points.

On objects, the object model is applied to the sparse input without change. On
scenes, the input distribution changes more, so we fine-tune the scene model of
Section~\ref{subsec:supervision_scene} on crops thinned at random spacings.
During fine-tuning, the encoder sees only the sparse points, and the target is
read from the dense scan. Queries between the sparse points therefore still
receive a target, and the shell target~\eqref{eq:shell} is applied around the
dense points. The query mixture and the schedule are given in the supplementary
material (Sec.~S5-D).

On objects, the field is read once at the sparse points. On scenes, the field is
also read at candidate positions around each sparse point. The candidates are 12
samples in the tangent disc of radius $0.75\,s$, offset along the estimated
normal by up to $0.25\,s$. We use this fixed set of candidates instead of moving
query points by gradient ascent on the field, so that the sparse-map experiment
of Section~\ref{subsec:exp_sparse} can compare the two methods fairly. In that
experiment, a continuous variant of Harris-3D is evaluated at exactly the same
positions as \ours{}, and the two methods differ only in how the response is
computed. On scenes, the response is averaged in rank over the input and three
random rotations of it, and the choice between the candidates and the sparse
points alone is made for each spacing on the development crops. Keypoints are
extracted as in Section~\ref{subsec:extract}.

\section{Experiments}
\label{sec:exp}

We first describe the evaluation protocol shared by all experiments
(Section~\ref{subsec:protocol}). We then compare \ours{} with 3D keypoint methods
that use other definitions of saliency, on objects
(Section~\ref{subsec:exp_shapenet}) and on real indoor scenes
(Section~\ref{subsec:exp_real}). Section~\ref{subsec:exp_sparse} tests whether
the advantage remains on sparse maps. Section~\ref{subsec:exp_d2net} evaluates with a 2D
detector that is not used in training. Finally, Section~\ref{subsec:exp_ablation}
examines the design choices of Section~\ref{sec:method}.

\subsection{Protocol}
\label{subsec:protocol}

\paragraph{Task}
We evaluate 3D keypoints on the task they are meant to serve, which is
image-based 6-DoF camera localization. A point cloud serves as the map, and a
reference image of it with a known pose is available. Each method selects $K$
3D keypoints on the point cloud. These keypoints are projected into the reference
image, and a descriptor is extracted at each projection. The descriptor comes
from an off-the-shelf 2D detector and descriptor, which is SuperPoint unless
stated otherwise. A
query image is matched against these descriptors, and the pose of the query is
estimated with EPnP~\cite{lepetit2009epnp} and RANSAC~\cite{fischler1981random}. \emph{Only the source of the $K$ 3D points changes
between methods}, and all later steps are identical. This protocol follows the
registration protocol of UKPGAN~\cite{you2022ukpgan}, with registration replaced
by localization. Descriptors are read at the projected location of each 3D
point, without snapping to the nearest 2D detection. The remaining settings of
the localization pipeline are listed in the supplementary material (Sec.~S5-E).

\paragraph{Metrics}
\emph{Localization success} is the fraction of query images whose pose is
recovered within a rotation threshold and a translation threshold. On objects,
the thresholds are $10^\circ$ and $0.05$ in normalized coordinates, where the
longest side of an object is $0.9$. On real scenes, the primary thresholds are
$5^\circ$ and $10$\,cm. We also report a strict setting of $5^\circ$ and
$5$\,cm, and a loose setting of $10^\circ$ and $25$\,cm. \emph{Hit rate} is the
fraction of the keypoints visible in the reference image whose projection lies
near a 2D detection in that image. The tolerance is $4$\,px on rendered objects
and $8$\,px on the real photographs at half resolution. Hit rate measures image
detectability directly, without the effects of matching and RANSAC. Differences
between methods are reported with paired bootstrap intervals. The bootstrap
resamples objects on ShapeNet and crops on real scenes, so that the frame pairs
of one crop are resampled together. An asterisk marks a $95\%$ interval that
excludes zero.

\paragraph{Difficulty axis}
We use the keypoint budget $K$ as the difficulty axis. With a small $K$, the map
keeps few points, and every selected point has to be re-detected and matched.
With a large $K$, the map is dense enough that the choice of points matters
little. On objects, we sweep
$K$ from $1024$ down to $8$. A real-scene crop holds $2$--$7\times10^5$ points, so on real scenes we report
the small budgets $K\in\{64,80,96\}$, which keep only a small fraction of them. We
average the success of each frame pair over the three budgets.

\paragraph{Baselines}
We compare with five learned 3D keypoint methods, which are
USIP~\cite{li2019usip}, UKPGAN~\cite{you2022ukpgan}, D3Feat~\cite{bai2020d3feat},
SC3K~\cite{zohaib2023sc3k} and Key-Grid~\cite{hou2024keygrid}. The learned
methods are retrained on our data. The only exception is UKPGAN, for which we use
the universal weights released by its authors. We also compare with the
classical ISS~\cite{zhong2009iss} and Harris-3D~\cite{sipiran2011harris}
detectors. Farthest-point sampling and uniform random sampling serve as floors.
Harris-3D is the strongest baseline in our experiments and is reported
throughout. It is run with its standard $k=32$ neighbourhood. On objects and on
scenes, we also tuned this neighbourhood on the test set in favour of the
baseline, and our margins hold against the best setting (supplementary
material, Sec.~S2-C). The training of each learned method was checked, and the
trained USIP selects keypoints that are far more repeatable under rotation than
ISS and random points (Sec.~S4). Deviations from the original settings of the
baselines are also listed there.

Two further arms on objects are not competitors and are marked as such. The arm
\texttt{b4} evaluates the baked target~\eqref{eq:aggregate} directly, so it is
the ceiling of what our supervision can give. The arm \texttt{bref} selects
points by back-projecting detections from the reference image. It uses
information from the test input and is therefore an oracle.

\paragraph{Datasets}
\emph{Objects.} We use 805 ShapeNet~\cite{chang2015shapenet} models from 11
categories, split by model into 739 training models and 66 held-out models.
Localization is evaluated with rendered query views on 60 held-out objects,
which are drawn from the 66 at random with a fixed seed. Both the reference and the query views are rendered without texture in the
geometry-only encoding~\eqref{eq:encoding}, and the descriptors are extracted
from these renderings.

\emph{Real scenes.} We use 321 crops from 37
SceneFun3D~\cite{delitzas2024scenefun3d} rooms, built from the Faro laser scans.
Each crop is the region co-visible from one cluster of real camera frames
(Section~\ref{subsec:supervision_scene}). The training and evaluation rooms are
disjoint. The model is trained on 291 crops from 31 rooms and evaluated on 30
crops from 6 other rooms. Six of the evaluation crops were used during
development to choose the model configuration. We exclude them, and all
real-scene numbers below are on the remaining 24 test crops. Each crop
contributes 10 real frames from its cluster, except one test crop with only six
frames. Ordered pairs of these frames serve as reference and query, which gives
2100 pairs per budget. The query images are
the original photographs at half resolution. Results on all 30 evaluation crops,
including the six development crops, are given in the supplementary material
(Sec.~S2-A). The sample list of every split and the content hashes of the
processed data are given in Sec.~S6.

\subsection{Objects}
\label{subsec:exp_shapenet}

\begin{table}[t]
\centering
\caption{Objects: hit rate (\%)}
\label{tab:shapenet_hit}
\footnotesize
\begin{threeparttable}
\begin{tabular*}{\columnwidth}{@{\extracolsep{\fill}}lccc@{}}
\toprule
Method & $K{=}32$ & $16$ & $8$\\
\midrule
\texttt{b4} (ceiling, baked target) & 61.6 & 65.5 & 70.0\\
\midrule
\textbf{\ours{}} & \textbf{43.6} & \textbf{52.1} & \textbf{61.3}\\
Harris-3D & 31.0 & 35.7 & 36.4\\
ISS & 18.4 & 21.1 & 27.3\\
FPS & 18.0 & 22.9 & 31.2\\
Key-Grid & 14.9 & 15.8 & 17.2\\
D3Feat & 14.3 & 14.8 & 15.1\\
Random & 12.8 & 13.4 & 14.0\\
UKPGAN & 11.1 & 10.3 & 12.5\\
SC3K & 11.4 & 11.0 & 10.8\\
USIP & 10.7 & 12.4 & 10.4\\
\bottomrule
\end{tabular*}
\begin{tablenotes}[para,flushleft]\footnotesize
\item[]\hspace{-\labelsep}The evaluation uses 60 held-out objects. \texttt{b4} reads the baked target
directly and is the ceiling of the supervision. The hit rate is averaged over
the pairs of each object and then over the objects. For each method and budget,
objects with no keypoint visible in the reference views are left out.
\end{tablenotes}
\end{threeparttable}
\end{table}

\begin{table}[t]
\centering
\caption{Objects: localization success rate (\%)}
\label{tab:shapenet_loc}
\footnotesize
\begin{threeparttable}
\begin{tabular*}{\columnwidth}{@{\extracolsep{\fill}}lccccc@{}}
\toprule
Method & $K{=}1024$ & $512$ & $256$ & $128$ & $64$\\
\midrule
\texttt{bref} (oracle) & 67.6 & 67.5 & 67.5 & 66.9 & 65.2\\
\texttt{b4} (ceiling) & 56.1 & 54.4 & 51.4 & 41.9 & 29.9\\
\midrule
\textbf{\ours{}} & \textbf{53.4} & \textbf{51.2} & \textbf{50.0} & \textbf{43.6} & \textbf{32.3}\\
Harris-3D & 50.9 & 48.3 & 40.4 & 29.8 & 18.0\\
ISS & 44.5 & 38.3 & 29.2 & 18.3 & 9.7\\
D3Feat & 44.5 & 34.9 & 24.0 & 14.5 & 6.3\\
FPS & 44.2 & 36.2 & 26.0 & 16.4 & 8.0\\
USIP & 39.6 & 26.5 & 11.6 & 5.2 & 2.6\\
Random & 39.3 & 30.8 & 20.6 & 12.8 & 5.5\\
UKPGAN & 26.0 & 15.6 & 7.3 & 3.9 & 1.5\\
\bottomrule
\end{tabular*}
\begin{tablenotes}[para,flushleft]\footnotesize
\item[]\hspace{-\labelsep}The evaluation uses 60 held-out objects. A pose counts as a success when the
rotation error is below $10^\circ$ and the translation error is below $0.05$ in
normalized coordinates. \texttt{bref} uses detections in the reference image and
is an oracle. \texttt{b4} is the ceiling of the supervision. SC3K and Key-Grid
output a fixed number of keypoints and are therefore not shown at these budgets
(supplementary material, Sec.~S4).
\end{tablenotes}
\end{threeparttable}
\end{table}

Tables~\ref{tab:shapenet_hit} and~\ref{tab:shapenet_loc} report the hit rate
and the localization success on objects. We make three observations from these
tables.

First, the margin over every competitor is significant at every budget on both
metrics, and it widens as the budget becomes smaller. Over Harris-3D, the margin
in hit rate is $+12.6$, $+16.4$ and $+24.6$ points at $K=32$, $16$ and $8$. The
margin in localization success grows from $+2.6$ points at $K=1024$ to $+9.6$,
$+13.9$ and $+14.3$ points at $K=256$, $128$ and $64$.

\begin{figure*}[t]
\centering
\includegraphics[width=\textwidth]{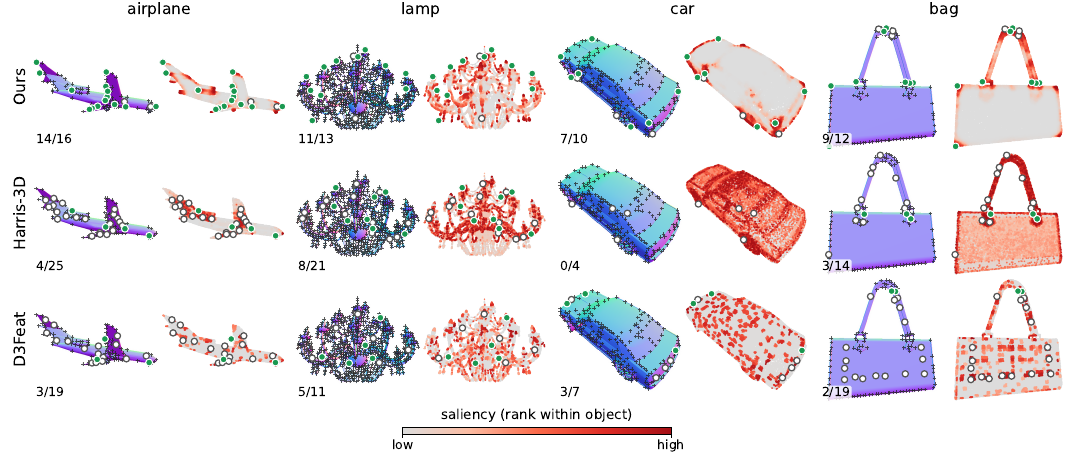}
\caption{\label{fig:qualitative}Keypoints of three methods on four held-out
objects. For each object, the left panel shows the keypoints in one reference
view, rendered as colour-coded normals for display. The right panel shows the
same keypoints on the visible points of the point cloud, coloured by the
saliency of each method. The saliency is shown as its rank within the object, so
that the three methods share one colour scale. Black crosses are the SuperPoint
detections in the reference view. A keypoint within $4$\,px of a detection is a
hit (green), and otherwise a miss (white). Each method selects $K{=}32$
keypoints, and the numbers give the hits over those of the $32$ keypoints that
are visible in this view. The objects were selected for illustration among
those with a large difference to Harris-3D.}
\end{figure*}

Second, the hit rates of the five learned 3D keypoint methods are within
$3.7$ points of uniform random sampling at every budget. Farthest-point sampling
is included as a floor that only spreads the keypoints evenly over the surface,
without any notion of saliency. Even this floor scores higher than all five
methods at every budget. This suggests that the learned methods select locations to which a
2D detector tends not to respond, such as semantic parts, symmetry centres or
locations where a 3D descriptor is distinctive. This gap is visible only because
the table includes the random and farthest-point floors. Fig.~\ref{fig:qualitative} illustrates
this difference. The saliency of \ours{} is concentrated on edges, corners and
thin parts, while the saliency of Harris-3D and D3Feat is spread over the
surfaces.

Third, at small budgets \ours{} is as good as its own supervision. At $K=128$
and $K=64$, the localization success of \ours{} is at the level of the baked
target \texttt{b4}. The differences are $+1.8$ and $+2.4$ points, and both
intervals contain zero. This holds although \ours{} sees only the point cloud,
while \texttt{b4} is computed from $128$ rendered views of the mesh.

Tuning the neighbourhood of Harris-3D on the test objects does not close the gap
(supplementary material, Sec.~S2-C).

\subsection{Real indoor scenes}
\label{subsec:exp_real}

We define saliency as image detectability, which is the response of a 2D
detector in real images. The setting closest to this definition is a real scene
queried with real photographs, and Table~\ref{tab:sf3d} reports the results in
this setting. The map is built from laser-scanned geometry, and the query images
are real photographs of a room that the model has never seen.

\begin{table}[t]
\centering
\caption{Real indoor scenes: localization success and hit rate (\%)}
\label{tab:sf3d}
\footnotesize
\begin{threeparttable}
\setlength{\tabcolsep}{4pt}
\begin{tabular*}{\columnwidth}{@{\extracolsep{\fill}}lcccc@{}}
\toprule
& \multicolumn{3}{c}{Success (\%)} & Hit\\
\cmidrule(lr){2-4}
Method & $5^\circ$,\,5\,cm & $5^\circ$,\,10\,cm & $10^\circ$,\,25\,cm & rate (\%)\\
\midrule
\textbf{\ours{}} & \textbf{37.7} & \textbf{59.7} & \textbf{75.3} & \textbf{42.1}\\
Harris-3D & 31.3 & 52.3 & 69.7 & 36.3\\
D3Feat & 10.0 & 23.1 & 40.0 & 20.7\\
SC3K & 8.9 & 21.9 & 40.6 & 20.3\\
ISS & 9.0 & 21.9 & 40.4 & 19.9\\
USIP & 8.5 & 21.8 & 40.2 & 18.5\\
Key-Grid & 6.9 & 18.7 & 36.3 & 19.0\\
UKPGAN & 4.7 & 13.9 & 31.0 & 20.7\\
Random & 7.4 & 18.2 & 36.0 & 19.2\\
\bottomrule
\end{tabular*}
\begin{tablenotes}[para,flushleft]\footnotesize
\item[]\hspace{-\labelsep}The evaluation uses 24 SceneFun3D test crops from rooms unseen in training and
2100 real frame pairs. Success is averaged over $K\in\{64,80,96\}$.
\end{tablenotes}
\end{threeparttable}
\end{table}

The margin over Harris-3D is $+6.4$, $+7.4$ and $+5.6$ points of success under
the strict, primary and loose thresholds, and $+5.8$ points of hit rate. Every
interval excludes zero. The margins over the other methods are also significant
in every column. The gain is not limited to crossing the thresholds. Over
all frame pairs at $K=80$, the geometric mean of the translation error of
\ours{} is $0.76\times$ that of Harris-3D, and the geometric mean of the
rotation error is $0.75\times$. Both ratios are significant, and the details of
this computation are given in the supplementary material (Sec.~S2-E).
Fig.~\ref{fig:sf3d_qual} shows two pairs in which this difference decides the
outcome.

Tuning the neighbourhood of Harris-3D on the test crops does not close the gap
either (Sec.~S2-C).

The learned 3D keypoint methods perform much worse here than on objects. All
five, and ISS, are within $5.1$ points of random sampling in every column,
and $37$ to $46$ points below \ours{} under the primary threshold. On real
scenes, the points they select are rarely re-detected in real photographs, which
the hit-rate column measures directly.

\begin{figure*}[t]
\centering
\includegraphics[width=\textwidth]{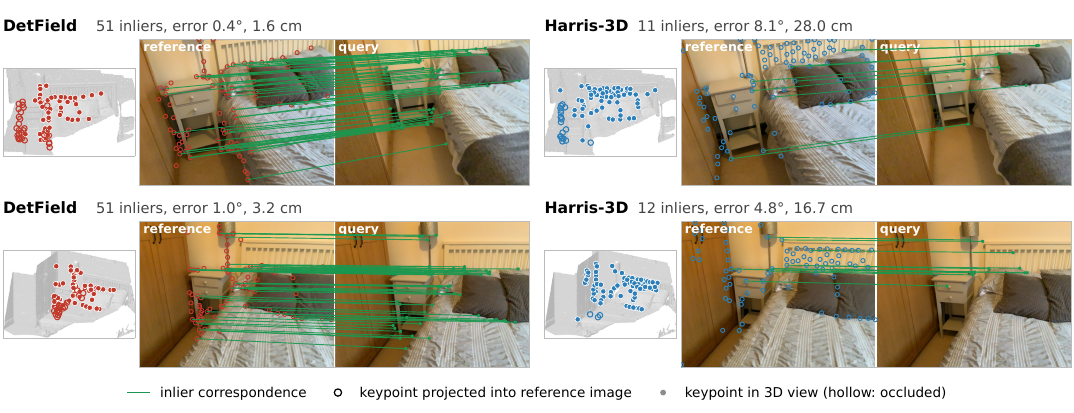}
\caption{\label{fig:sf3d_qual}Localization on two held-out real pairs at
$K{=}80$, with \ours{} on the left and Harris-3D on the right. In both pairs,
\ours{} recovers the pose within the primary threshold ($5^\circ$, $10$\,cm),
and Harris-3D does not. The Harris-3D keypoints gather on the repetitive slats
of the headboard. The pairs were selected to illustrate the difference.}
\end{figure*}

\subsection{Sparse maps}
\label{subsec:exp_sparse}

The maps in the previous experiments are dense point clouds. We now thin them by
voxel subsampling as described in Section~\ref{subsec:sparse}, and repeat the
localization benchmark with the same budgets and thresholds. On a sparse map,
Harris-3D can only rank the stored points, and its neighbourhood statistics are
computed from fewer points. We therefore give Harris-3D two advantages. First,
its neighbourhood size $k\in\{8,16,32\}$ is chosen separately for each spacing.
On objects, we report the best $k$ on the test objects, which favours the
baseline. On scenes, $k$ was fixed on development crops before the test set was
built.
Second, we add a continuous variant of Harris-3D. This variant fits the same
local quadric to the $k$ nearest sparse points, and evaluates the Harris-3D
response at the same positions as \ours{}. These are the sparse points on
objects and the candidate positions of Section~\ref{subsec:sparse} on scenes.
The two methods then differ only in how the response is computed.

\begin{figure}[t]
\centering
\includegraphics[width=\columnwidth]{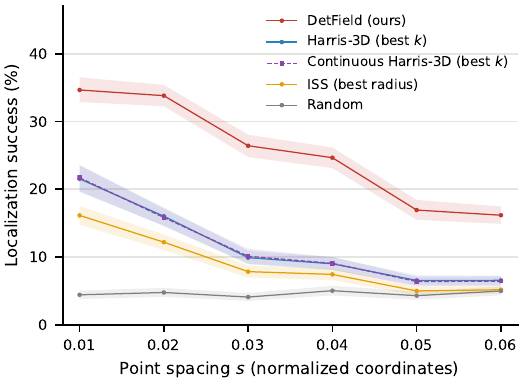}
\caption{\label{fig:sparse_trend}Localization success on sparse objects as the
point spacing $s$ grows, on 130 held-out ShapeNet models and averaged over
$K\in\{32,64,128,256\}$. Shaded bands are $95\%$ bootstrap intervals over
objects. Harris-3D and its continuous variant overlap almost exactly.}
\end{figure}

\paragraph{Objects}
For sparse objects, we use 130 further ShapeNet models from 15 categories with
planar parts. These models are disjoint from the 805 models above. The queries
are rendered as in Section~\ref{subsec:exp_shapenet}, and success is measured
within $10^\circ$ and $0.05$ and averaged over $K\in\{32,64,128,256\}$. The
spacing $s$ runs from $0.01$ to $0.06$ in normalized coordinates.
Fig.~\ref{fig:sparse_trend} shows that \ours{} is ahead of every baseline at
every spacing, and the intervals do not overlap. As the map becomes sparser,
Harris-3D and ISS fall towards random sampling, while \ours{} keeps a large part
of its success. The continuous variant of Harris-3D performs almost the same as
the original. Evaluating the classical response between the samples therefore
does not recover what the samples have lost. The numbers are listed in the
supplementary material (Table~S2).

\paragraph{Real scenes}
For real scenes, we report spacings of $5$ and $10$\,cm, which are typical voxel
sizes of stored localization maps. The test set for this
experiment is new. It contains 64 crops from 8 SceneFun3D rooms, which were used
neither for training nor for any development decision. The crops are built with
the procedure of Section~\ref{subsec:supervision_scene}.
Table~\ref{tab:sparse_sf3d} reports the success under the primary and the strict
thresholds. At both spacings, the margin of \ours{} over both Harris-3D variants
is significant under both thresholds. This holds whether the bootstrap resamples
crops or rooms.

\begin{table}[t]
\centering
\caption{Sparse maps: localization success (\%) on real indoor scenes}
\label{tab:sparse_sf3d}
\footnotesize
\begin{threeparttable}
\setlength{\tabcolsep}{4pt}
\begin{tabular*}{\columnwidth}{@{\extracolsep{\fill}}lcccc@{}}
\toprule
& \multicolumn{2}{c}{$5^\circ$,\,10\,cm} & \multicolumn{2}{c}{$5^\circ$,\,5\,cm}\\
\cmidrule(lr){2-3}\cmidrule(lr){4-5}
Method & $s{=}5$\,cm & $s{=}10$\,cm & $s{=}5$\,cm & $s{=}10$\,cm\\
\midrule
\textbf{\ours{}} & $\mathbf{32.7}^{*}$ & $\mathbf{25.8}^{*}$ & $\mathbf{18.9}^{*}$ & $\mathbf{13.7}^{*}$\\
Harris-3D & 29.0 & 22.5 & 16.1 & 11.5\\
Continuous Harris-3D & 28.6 & 23.0 & 16.0 & 11.7\\
\bottomrule
\end{tabular*}
\begin{tablenotes}[para,flushleft]\footnotesize
\item[]\hspace{-\labelsep}The evaluation uses 64 crops from 8 SceneFun3D rooms unseen in training and
development. Success is averaged over $K\in\{64,80,96\}$, and $s$ is the point
spacing.
\item[$*$] The margins over both Harris-3D variants have $95\%$ intervals that
exclude zero (paired bootstrap over crops).
\end{tablenotes}
\end{threeparttable}
\end{table}

The supplementary material (Sec.~S3) further shows that a corner head added to
\ours{} can place keypoints on object corners that fall between the samples.

\subsection{Evaluation with an unseen 2D detector}
\label{subsec:exp_d2net}

The supervision of \ours{} comes from SuperPoint, so an evaluation with SuperPoint
tests the model against its own teacher. A stronger test uses a 2D detector that
\ours{} has never seen. We repeat the full object pipeline with
D2-Net~\cite{dusmanu2019d2}. D2-Net detects and describes jointly from a dense
feature map, so it supports the same protocol without snapping. The 3D keypoints of
every method are unchanged, and only the 2D detector and descriptor used for
matching are replaced.

\begin{table}[t]
\centering
\caption{Unseen 2D detector: localization success rate (\%) with D2-Net}
\label{tab:d2net}
\footnotesize
\begin{threeparttable}
\begin{tabular*}{\columnwidth}{@{\extracolsep{\fill}}lccccc@{}}
\toprule
Method & $K{=}1024$ & $512$ & $256$ & $128$ & $64$\\
\midrule
\texttt{bref} (oracle) & 47.5 & 47.6 & 46.8 & 40.3 & 27.7\\
\texttt{b4} (SuperPoint-baked) & 43.4 & 40.6 & 33.6 & 23.9 & 12.8\\
\midrule
\textbf{\ours{}} & \textbf{43.0} & \textbf{40.6} & \textbf{38.5} & \textbf{32.9} & \textbf{24.0}\\
Harris-3D & 41.1 & 36.9 & 30.3 & 19.6 & 13.8\\
ISS & 40.1 & 33.5 & 26.8 & 18.9 & 10.2\\
FPS & 40.1 & 33.4 & 24.8 & 16.3 & 9.1\\
D3Feat & 39.8 & 32.7 & 23.9 & 15.7 & 7.9\\
USIP & 39.1 & 28.4 & 13.8 & 7.2 & 3.9\\
Random & 35.1 & 29.5 & 23.3 & 14.3 & 8.1\\
UKPGAN & 24.6 & 15.7 & 7.7 & 3.3 & 1.5\\
\bottomrule
\end{tabular*}
\begin{tablenotes}[para,flushleft]\footnotesize
\item[]\hspace{-\labelsep}D2-Net serves as the 2D detector and descriptor at test time and is not used
in training. The evaluation uses 60 held-out objects. \texttt{bref} and
\texttt{b4} are defined as in Table~\ref{tab:shapenet_loc}, and \texttt{b4} is
baked with SuperPoint.
\end{tablenotes}
\end{threeparttable}
\end{table}

The advantage carries over to D2-Net. Against the seven competitors, 34 of the
35 paired comparisons in Table~\ref{tab:d2net} are significant. The exception is
Harris-3D at $K=1024$, where the map is dense enough for most selections to work.
The only arm that significantly outperforms \ours{} is the oracle \texttt{bref}.

The most informative row is \texttt{b4}. It is built by baking SuperPoint
responses, so under SuperPoint it is an upper bound on what our supervision can
give. Under D2-Net it is no longer an upper bound, and \ours{} outperforms it by
$+4.9$, $+9.1$ and $+11.2$ points at $K{=}256$, $128$ and $64$, all significant.
This result is unlikely if \ours{} had only memorized where SuperPoint responds.
It indicates that \ours{} has learned a more general notion of which surface
structures a local 2D detector responds to, which supports the main claim of
this paper.

The same comparison also answers a direct objection to learning a field. If
saliency is defined by the response of a 2D detector, one could simply store the
back-projected responses, which is what \texttt{b4} does. At small budgets,
\ours{} matches \texttt{b4} under the 2D detector used for supervision
(Table~\ref{tab:shapenet_loc}), and outperforms it under a 2D detector that was
not used. In both cases, \ours{} requires only the point cloud, while
\texttt{b4} requires the mesh and its renderings.

\subsection{Ablations}
\label{subsec:exp_ablation}

Table~\ref{tab:ablation} changes one design choice at a time on the real-scene
benchmark, where the margins are largest. All variants, including the full
model, are trained in the same way. They are trained for $80$ epochs from the
same initialization, which is half the schedule of the model in
Table~\ref{tab:sf3d}, and the last epoch is evaluated. Every variant that
removes or replaces a part of the method scores below the full model, and the
larger losses are significant. The only addition that we tested is a Harris-3D
prior in the target, and it does not change the result. The last row gives the
Harris-3D baseline under the same evaluation as a reference. The groups of the
table follow the paragraphs below.

\begin{table}[t]
\centering
\caption{Ablation: localization success (\%) on real indoor scenes}
\label{tab:ablation}
\footnotesize
\begin{threeparttable}
\setlength{\tabcolsep}{4pt}
\begin{tabular*}{\columnwidth}{@{\extracolsep{\fill}}lrr@{}}
\toprule
Variant & Success & $\Delta$\\
\midrule
\textbf{Full model} & \textbf{58.2} & ---\\
\midrule
\multicolumn{3}{@{}l}{\emph{Per-query features}}\\
\quad features to the encoder only, no skip path & 54.2 & $-4.0^{*}$\\
\quad no geometric features & 39.6 & $-18.6^{*}$\\
\midrule
\multicolumn{3}{@{}l}{\emph{Invariants or Harris-3D responses}}\\
\quad Harris-3D responses instead of invariants & 56.4 & $-1.7$\\
\quad \ldots\ and rendered target & 52.9 & $-5.3^{*}$\\
\midrule
\multicolumn{3}{@{}l}{\emph{Photographic target}}\\
\quad rendered target instead & 51.1 & $-7.1^{*}$\\
\quad plus a Harris-3D prior & 58.8 & $+0.7$\\
\midrule
\multicolumn{3}{@{}l}{\emph{Test-time averaging} (Sec.~S2-D)}\\
\quad single pass, no averaging & 57.2 & $-1.0$\\
\midrule
\emph{Reference:} Harris-3D baseline & 52.3 & $-5.9^{*}$\\
\bottomrule
\end{tabular*}
\begin{tablenotes}[para,flushleft]\footnotesize
\item[]\hspace{-\labelsep}The evaluation uses the 24 real-scene test crops under the primary threshold
($5^\circ$, $10$\,cm). Each row changes one choice of the full model. The full
model and all variants are trained for 80 epochs. $\Delta$ is the paired
difference to the full model (bootstrap over crops).
\item[$*$] The $95\%$ interval excludes zero.
\end{tablenotes}
\end{threeparttable}
\end{table}

\paragraph{Per-query features}
Without any geometric features, the model reaches only $39.6\%$ ($-18.6$
points). Giving the nine features to the encoder alone recovers most of this
loss, but still costs $4.0$ points. The model is then level with Harris-3D, with
a difference of $+1.9$ points and an interval that contains zero. The variation
of the features inside a voxel does not survive the pooling
(Section~\ref{subsec:qfeat}), and only the per-query path delivers it to the
decoder. Through the encoder, the features add $14.6$ points, and the skip path
adds a further $4.0$ points.

\paragraph{Invariants or Harris-3D responses}
Feeding the invariants from which Harris-3D is computed scores at least as well
as feeding the Harris-3D responses at the same three scales. The variant with
Harris-3D responses is $1.7$ points lower, and the difference is not
significant. Together with the rendered target, the Harris-3D input loses $5.3$
points. The network therefore does better when it receives the ingredients of
the corner measure than when it receives the measure itself. With this choice,
the input of the model also contains no Harris-3D component.

\paragraph{The photographic target}
Replacing the 2D detector response measured on real photographs with the
response rendered from geometry costs $7.1$ points. This is the largest loss
after removing the features altogether. The supervision therefore works better
when it is measured in the kind of image that the map will be queried with.
Averaging the photographic target in rank with the Harris-3D response, as a
curvature prior, changes the result by $+0.7$ points, and the interval contains
zero. We therefore keep Harris-3D out of the supervision. Harris-3D then enters
neither the input nor the target of the model that it is compared against.

Section~S2-D of the supplementary material gives two further ablations. They
concern the test-time averaging and the per-query features on objects.

\section{Limitations}
\label{sec:limits}

\paragraph{Dependence on the 2D detector}
\ours{} learns the response of the 2D detector used for supervision, so it also
inherits the limits of this detector. For example, the 2D detector is not
equivariant to mirroring, and the construction symmetrizes its response for this
reason. On objects, the supervision is computed from renderings, so it can also
reflect how a mesh is tessellated rather than the shape itself. The supervision
on real scenes is measured in photographs and does not have this issue.

\paragraph{Scope}
The supervision is built from a single 2D detector. Training with other 2D detectors and evaluating on outdoor scenes
are left for future work.

\section{Conclusion}
\label{sec:conclusion}

In this paper, we defined 3D saliency as image detectability, which is whether a
2D detector will detect a surface point again in an image. This property decides
whether a camera can be localized against a 3D map. We proposed \ours{}, which
learns this property as a continuous field from geometry alone. \ours{} is
supervised by the responses of a 2D detector on renderings of objects and on real
photographs of scenes. At every query point, it also receives the curvature
invariants from which Harris-3D is computed, through a path that bypasses the
voxel pooling of the encoder.

In a localization benchmark that changes only the source of the 3D points,
\ours{} outperforms Harris-3D and five learned 3D keypoint methods on objects and
on real indoor scenes. On objects, it also outperforms its own training target
under a 2D detector that is not used in training. On sparse maps, it stays ahead
of Harris-3D. Image detectability can be checked against an independent 2D
detector, as the evaluation with D2-Net shows. The same definition can also
serve as a training objective for other 3D keypoint detectors built for
localization.

\bibliographystyle{IEEEtran}
\bibliography{biblo,c1_extra}

\begin{thebibliography}{10}
\providecommand{\url}[1]{#1}
\csname url@samestyle\endcsname
\providecommand{\newblock}{\relax}
\providecommand{\bibinfo}[2]{#2}
\providecommand{\BIBentrySTDinterwordspacing}{\spaceskip=0pt\relax}
\providecommand{\BIBentryALTinterwordstretchfactor}{4}
\providecommand{\BIBentryALTinterwordspacing}{\spaceskip=\fontdimen2\font plus
\BIBentryALTinterwordstretchfactor\fontdimen3\font minus
  \fontdimen4\font\relax}
\providecommand{\BIBforeignlanguage}[2]{{%
\expandafter\ifx\csname l@#1\endcsname\relax
\typeout{** WARNING: IEEEtran.bst: No hyphenation pattern has been}%
\typeout{** loaded for the language `#1'. Using the pattern for}%
\typeout{** the default language instead.}%
\else
\language=\csname l@#1\endcsname
\fi
#2}}
\providecommand{\BIBdecl}{\relax}
\BIBdecl

\bibitem{you2020keypointnet}
Y.~You, Y.~Lou, C.~Li, Z.~Cheng, L.~Li, L.~Ma, C.~Lu, and W.~Wang,
  ``{KeypointNet}: A large-scale {3D} keypoint dataset aggregated from numerous
  human annotations,'' in \emph{IEEE/CVF Conference on Computer Vision and
  Pattern Recognition (CVPR)}, 2020, pp. 13\,647--13\,656.

\bibitem{li2019usip}
J.~Li and G.~Lee, ``{USIP}: Unsupervised stable interest point detection from
  {3D} point clouds,'' in \emph{IEEE/CVF International Conference on Computer
  Vision (ICCV)}, 2019, pp. 361--370.

\bibitem{you2022ukpgan}
Y.~You, W.~Liu, Y.~Ze, Y.~Li, W.~Wang, and C.~Lu, ``{UKPGAN}: A general
  self-supervised keypoint detector,'' in \emph{IEEE/CVF Conference on Computer
  Vision and Pattern Recognition (CVPR)}, 2022, pp. 17\,042--17\,051.

\bibitem{bai2020d3feat}
X.~Bai, Z.~Luo, L.~Zhou, H.~Fu, L.~Quan, and C.~Tai, ``{D3Feat}: Joint learning
  of dense detection and description of {3D} local features,'' in
  \emph{IEEE/CVF Conference on Computer Vision and Pattern Recognition (CVPR)},
  2020, pp. 6359--6367.

\bibitem{feng20192d3d}
M.~Feng, S.~Hu, M.~H. Ang, and G.~H. Lee, ``{2D3D-MatchNet}: Learning to match
  keypoints across {2D} image and {3D} point cloud,'' in \emph{IEEE
  International Conference on Robotics and Automation (ICRA)}.\hskip 1em plus
  0.5em minus 0.4em\relax IEEE, 2019, pp. 4790--4796.

\bibitem{wang2021p2}
B.~Wang, C.~Chen, Z.~Cui, J.~Qin, C.~X. Lu, Z.~Yu, P.~Zhao, Z.~Dong, F.~Zhu,
  N.~Trigoni \emph{et~al.}, ``{P2-Net}: Joint description and detection of
  local features for pixel and point matching,'' in \emph{IEEE/CVF
  International Conference on Computer Vision (ICCV)}, 2021, pp.
  16\,004--16\,013.

\bibitem{detone2018superpoint}
D.~DeTone, T.~Malisiewicz, and A.~Rabinovich, ``{SuperPoint}: Self-supervised
  interest point detection and description,'' in \emph{IEEE Conference on
  Computer Vision and Pattern Recognition Workshops (CVPRW)}, 2018, pp.
  224--236.

\bibitem{qi2017pointnet}
C.~Qi, H.~Su, K.~Mo, and L.~Guibas, ``{PointNet}: Deep learning on point sets
  for {3D} classification and segmentation,'' in \emph{IEEE Conference on
  Computer Vision and Pattern Recognition (CVPR)}, 2017, pp. 652--660.

\bibitem{qi2017pointnet++}
C.~R. Qi, L.~Yi, H.~Su, and L.~J. Guibas, ``{PointNet++}: Deep hierarchical
  feature learning on point sets in a metric space,'' in \emph{Advances in
  Neural Information Processing Systems (NeurIPS)}, vol.~30, 2017.

\bibitem{wang2019dynamic}
Y.~Wang, Y.~Sun, Z.~Liu, S.~E. Sarma, M.~M. Bronstein, and J.~M. Solomon,
  ``Dynamic graph {CNN} for learning on point clouds,'' \emph{ACM Transactions
  on Graphics}, vol.~38, no.~5, pp. 1--12, 2019.

\bibitem{jin2024keypointdetr}
H.~Jin, Y.~Shen, J.~Lou, K.~Zhou, and Y.~Zheng, ``{KeypointDETR}: An end-to-end
  {3D} keypoint detector,'' in \emph{European Conference on Computer Vision
  (ECCV)}, 2024, pp. 374--390.

\bibitem{back2024to3d}
T.~Wimmer, P.~Wonka, and M.~Ovsjanikov, ``Back to {3D}: Few-shot {3D} keypoint
  detection with back-projected {2D} features,'' in \emph{IEEE/CVF Conference
  on Computer Vision and Pattern Recognition (CVPR)}, 2024, pp. 4154--4164.

\bibitem{zohaib2023sc3k}
M.~Zohaib and A.~Del~Bue, ``{SC3K}: Self-supervised and coherent {3D} keypoints
  estimation from rotated, noisy, and decimated point cloud data,'' in
  \emph{IEEE/CVF International Conference on Computer Vision (ICCV)}, 2023, pp.
  22\,509--22\,519.

\bibitem{hou2024keygrid}
C.~Hou, Z.~Xue, B.~Zhou, J.~Ke, L.~Shao, and H.~Xu, ``{Key-Grid}: Unsupervised
  {3D} keypoints detection using grid heatmap features,'' in \emph{Advances in
  Neural Information Processing Systems (NeurIPS)}, vol.~37, 2024, pp.
  49\,154--49\,179.

\bibitem{zhong2009iss}
Y.~Zhong, ``Intrinsic shape signatures: A shape descriptor for {3D} object
  recognition,'' in \emph{IEEE International Conference on Computer Vision
  Workshops (ICCVW)}, 2009, pp. 689--696.

\bibitem{sipiran2011harris}
I.~Sipiran and B.~Bustos, ``{Harris} {3D}: A robust extension of the {Harris}
  operator for interest point detection on {3D} meshes,'' \emph{The Visual
  Computer}, vol.~27, pp. 963--976, 2011.

\bibitem{steder2010narf}
B.~Steder, R.~B. Rusu, K.~Konolige, and W.~Burgard, ``Point feature extraction
  on {3D} range scans taking into account object boundaries,'' in \emph{IEEE
  International Conference on Robotics and Automation (ICRA)}, 2011, pp.
  2601--2608.

\bibitem{yew20183dfeat}
Z.~J. Yew and G.~H. Lee, ``{3DFeat-Net}: Weakly supervised local {3D} features
  for point cloud registration,'' in \emph{European Conference on Computer
  Vision (ECCV)}, 2018, pp. 630--646.

\bibitem{huang2021predator}
S.~Huang, Z.~Gojcic, M.~Usvyatsov, A.~Wieser, and K.~Schindler, ``{PREDATOR}:
  Registration of {3D} point clouds with low overlap,'' in \emph{IEEE/CVF
  Conference on Computer Vision and Pattern Recognition (CVPR)}, 2021, pp.
  4267--4276.

\bibitem{song2021mesh}
R.~Song, W.~Zhang, Y.~Zhao, Y.~Liu, and P.~L. Rosin, ``Mesh saliency: An
  independent perceptual measure or a derivative of image saliency?'' in
  \emph{IEEE/CVF Conference on Computer Vision and Pattern Recognition (CVPR)},
  2021, pp. 8853--8862.

\bibitem{kalogerakis2017projective}
E.~Kalogerakis, M.~Averkiou, S.~Maji, and S.~Chaudhuri, ``{3D} shape
  segmentation with projective convolutional networks,'' in \emph{IEEE
  Conference on Computer Vision and Pattern Recognition (CVPR)}, 2017, pp.
  3779--3788.

\bibitem{dutt2024diff3f}
N.~S. Dutt, S.~Muralikrishnan, and N.~J. Mitra, ``Diffusion {3D} features
  ({Diff3F}): Decorating untextured shapes with distilled semantic features,''
  in \emph{IEEE/CVF Conference on Computer Vision and Pattern Recognition
  (CVPR)}, 2024, pp. 4494--4504.

\bibitem{peng2020convolutional}
S.~Peng, M.~Niemeyer, L.~Mescheder, M.~Pollefeys, and A.~Geiger,
  ``Convolutional occupancy networks,'' in \emph{European Conference on
  Computer Vision (ECCV)}, 2020, pp. 523--540.

\bibitem{li2022snake}
C.~Zhong, P.~You, X.~Chen, H.~Zhao, F.~Sun, G.~Zhou, X.~Mu, C.~Gan, and
  W.~Huang, ``{SNAKE}: Shape-aware neural {3D} keypoint field,'' in
  \emph{Advances in Neural Information Processing Systems (NeurIPS)}, vol.~35,
  2022, pp. 7052--7064.

\bibitem{cciccek20163dunet}
{\"O}.~{\c{C}}i{\c{c}}ek, A.~Abdulkadir, S.~Lienkamp, T.~Brox, and
  O.~Ronneberger, ``{3D} {U-Net}: learning dense volumetric segmentation from
  sparse annotation,'' in \emph{International Conference on Medical Image
  Computing and Computer-Assisted Intervention (MICCAI)}, 2016, pp. 424--432.

\bibitem{delitzas2024scenefun3d}
A.~Delitzas, A.~Takmaz, F.~Tombari, R.~Sumner, M.~Pollefeys, and F.~Engelmann,
  ``{SceneFun3D}: Fine-grained functionality and affordance understanding in
  {3D} scenes,'' in \emph{IEEE/CVF Conference on Computer Vision and Pattern
  Recognition (CVPR)}, 2024, pp. 14\,531--14\,542.

\bibitem{koenderink1992surface}
J.~J. Koenderink and A.~J. van Doorn, ``Surface shape and curvature scales,''
  \emph{Image and Vision Computing}, vol.~10, no.~8, pp. 557--564, 1992.

\bibitem{lepetit2009epnp}
V.~Lepetit, F.~Moreno-Noguer, and P.~Fua, ``{EPnP}: An accurate {$O(n)$}
  solution to the {PnP} problem,'' \emph{International Journal of Computer
  Vision}, vol.~81, no.~2, pp. 155--166, 2009.

\bibitem{fischler1981random}
M.~A. Fischler and R.~C. Bolles, ``Random sample consensus: a paradigm for
  model fitting with applications to image analysis and automated
  cartography,'' \emph{Communications of the ACM}, vol.~24, no.~6, pp.
  381--395, 1981.

\bibitem{chang2015shapenet}
A.~Chang, T.~Funkhouser, L.~Guibas, P.~Hanrahan, Q.~Huang, Z.~Li, S.~Savarese,
  M.~Savva, S.~Song, H.~Su, J.~Xiao, L.~Yi, and F.~Yu, ``{ShapeNet}: An
  information-rich {3D} model repository,'' \emph{arXiv preprint
  arXiv:1512.03012}, 2015.

\bibitem{dusmanu2019d2}
M.~Dusmanu, I.~Rocco, T.~Pajdla, M.~Pollefeys, J.~Sivic, A.~Torii, and
  T.~Sattler, ``{D2-Net}: A trainable {CNN} for joint description and detection
  of local features,'' in \emph{IEEE/CVF Conference on Computer Vision and
  Pattern Recognition (CVPR)}, 2019, pp. 8092--8101.

\end{thebibliography}


\begin{thebibliography}{10}
\providecommand{\url}[1]{#1}
\csname url@samestyle\endcsname
\providecommand{\newblock}{\relax}
\providecommand{\bibinfo}[2]{#2}
\providecommand{\BIBentrySTDinterwordspacing}{\spaceskip=0pt\relax}
\providecommand{\BIBentryALTinterwordstretchfactor}{4}
\providecommand{\BIBentryALTinterwordspacing}{\spaceskip=\fontdimen2\font plus
\BIBentryALTinterwordstretchfactor\fontdimen3\font minus
  \fontdimen4\font\relax}
\providecommand{\BIBforeignlanguage}[2]{{%
\expandafter\ifx\csname l@#1\endcsname\relax
\typeout{** WARNING: IEEEtran.bst: No hyphenation pattern has been}%
\typeout{** loaded for the language `#1'. Using the pattern for}%
\typeout{** the default language instead.}%
\else
\language=\csname l@#1\endcsname
\fi
#2}}
\providecommand{\BIBdecl}{\relax}
\BIBdecl

\bibitem{koenderink1992surface}
J.~J. Koenderink and A.~J. van Doorn, ``Surface shape and curvature scales,''
  \emph{Image and Vision Computing}, vol.~10, no.~8, pp. 557--564, 1992.

\bibitem{chang2015shapenet}
A.~Chang, T.~Funkhouser, L.~Guibas, P.~Hanrahan, Q.~Huang, Z.~Li, S.~Savarese,
  M.~Savva, S.~Song, H.~Su, J.~Xiao, L.~Yi, and F.~Yu, ``{ShapeNet}: An
  information-rich {3D} model repository,'' \emph{arXiv preprint
  arXiv:1512.03012}, 2015.

\bibitem{detone2018superpoint}
D.~DeTone, T.~Malisiewicz, and A.~Rabinovich, ``{SuperPoint}: Self-supervised
  interest point detection and description,'' in \emph{IEEE Conference on
  Computer Vision and Pattern Recognition Workshops (CVPRW)}, 2018, pp.
  224--236.

\bibitem{sipiran2011harris}
I.~Sipiran and B.~Bustos, ``{Harris} {3D}: A robust extension of the {Harris}
  operator for interest point detection on {3D} meshes,'' \emph{The Visual
  Computer}, vol.~27, pp. 963--976, 2011.

\bibitem{bai2020d3feat}
X.~Bai, Z.~Luo, L.~Zhou, H.~Fu, L.~Quan, and C.~Tai, ``{D3Feat}: Joint learning
  of dense detection and description of {3D} local features,'' in
  \emph{IEEE/CVF Conference on Computer Vision and Pattern Recognition (CVPR)},
  2020, pp. 6359--6367.

\bibitem{zhong2009iss}
Y.~Zhong, ``Intrinsic shape signatures: A shape descriptor for {3D} object
  recognition,'' in \emph{IEEE International Conference on Computer Vision
  Workshops (ICCVW)}, 2009, pp. 689--696.

\bibitem{zohaib2023sc3k}
M.~Zohaib and A.~Del~Bue, ``{SC3K}: Self-supervised and coherent {3D} keypoints
  estimation from rotated, noisy, and decimated point cloud data,'' in
  \emph{IEEE/CVF International Conference on Computer Vision (ICCV)}, 2023, pp.
  22\,509--22\,519.

\bibitem{li2019usip}
J.~Li and G.~Lee, ``{USIP}: Unsupervised stable interest point detection from
  {3D} point clouds,'' in \emph{IEEE/CVF International Conference on Computer
  Vision (ICCV)}, 2019, pp. 361--370.

\bibitem{hou2024keygrid}
C.~Hou, Z.~Xue, B.~Zhou, J.~Ke, L.~Shao, and H.~Xu, ``{Key-Grid}: Unsupervised
  {3D} keypoints detection using grid heatmap features,'' in \emph{Advances in
  Neural Information Processing Systems (NeurIPS)}, vol.~37, 2024, pp.
  49\,154--49\,179.

\bibitem{you2022ukpgan}
Y.~You, W.~Liu, Y.~Ze, Y.~Li, W.~Wang, and C.~Lu, ``{UKPGAN}: A general
  self-supervised keypoint detector,'' in \emph{IEEE/CVF Conference on Computer
  Vision and Pattern Recognition (CVPR)}, 2022, pp. 17\,042--17\,051.

\bibitem{delitzas2024scenefun3d}
A.~Delitzas, A.~Takmaz, F.~Tombari, R.~Sumner, M.~Pollefeys, and F.~Engelmann,
  ``{SceneFun3D}: Fine-grained functionality and affordance understanding in
  {3D} scenes,'' in \emph{IEEE/CVF Conference on Computer Vision and Pattern
  Recognition (CVPR)}, 2024, pp. 14\,531--14\,542.

\bibitem{fischler1981random}
M.~A. Fischler and R.~C. Bolles, ``Random sample consensus: a paradigm for
  model fitting with applications to image analysis and automated
  cartography,'' \emph{Communications of the ACM}, vol.~24, no.~6, pp.
  381--395, 1981.

\bibitem{wu20153d}
Z.~Wu, S.~Song, A.~Khosla, F.~Yu, L.~Zhang, X.~Tang, and J.~Xiao, ``{3D}
  {ShapeNets}: A deep representation for volumetric shapes,'' in \emph{IEEE
  Conference on Computer Vision and Pattern Recognition (CVPR)}, 2015, pp.
  1912--1920.

\bibitem{zhou2018open3d}
Q.-Y. Zhou, J.~Park, and V.~Koltun, ``{Open3D}: A modern library for {3D} data
  processing,'' \emph{arXiv preprint arXiv:1801.09847}, 2018.

\bibitem{zeng20173dmatch}
A.~Zeng, S.~Song, M.~Nie{\ss}ner, M.~Fisher, J.~Xiao, and T.~Funkhouser,
  ``{3DMatch}: Learning local geometric descriptors from {RGB-D}
  reconstructions,'' in \emph{IEEE Conference on Computer Vision and Pattern
  Recognition (CVPR)}, 2017, pp. 1802--1811.

\bibitem{cciccek20163dunet}
{\"O}.~{\c{C}}i{\c{c}}ek, A.~Abdulkadir, S.~Lienkamp, T.~Brox, and
  O.~Ronneberger, ``{3D} {U-Net}: learning dense volumetric segmentation from
  sparse annotation,'' in \emph{International Conference on Medical Image
  Computing and Computer-Assisted Intervention (MICCAI)}, 2016, pp. 424--432.

\bibitem{kazhdan2013screened}
M.~Kazhdan and H.~Hoppe, ``Screened {Poisson} surface reconstruction,''
  \emph{ACM Transactions on Graphics}, vol.~32, no.~3, pp. 29:1--29:13, 2013.

\bibitem{loshchilov2019adamw}
I.~Loshchilov and F.~Hutter, ``Decoupled weight decay regularization,'' in
  \emph{International Conference on Learning Representations (ICLR)}, 2019.

\bibitem{lowe2004distinctive}
D.~G. Lowe, ``Distinctive image features from scale-invariant keypoints,''
  \emph{International Journal of Computer Vision}, vol.~60, no.~2, pp. 91--110,
  2004.

\bibitem{lepetit2009epnp}
V.~Lepetit, F.~Moreno-Noguer, and P.~Fua, ``{EPnP}: An accurate {$O(n)$}
  solution to the {PnP} problem,'' \emph{International Journal of Computer
  Vision}, vol.~81, no.~2, pp. 155--166, 2009.

\end{thebibliography}

\end{document}


\title{Supplementary Material for\\
Detectable by Design: Learning a Continuous 3D Saliency Field\\
from 2D Detector Responses}

\author{Han Sun, Jun Li and Christy Jie Liang}

\maketitle

Section~\ref{sec:s_geometry} gives the geometric analysis that the main paper
relies on. Sections~\ref{sec:s_results} to~\ref{sec:s_sanity} give additional
results, the corner experiment and the checks of the baselines.
Sections~\ref{sec:s_impl} and~\ref{sec:s_repro} give the implementation details
and the data needed to reproduce the experiments. References to sections,
tables, figures and equations without the prefix~S are to the main paper.

\providecommand{\dN}{\mathrm{d}N}
\providecommand{\tr}{\operatorname{tr}}

\ifx\theoremstyle\undefined\else\theoremstyle{plain}\fi
\ifx\thmC\undefined
  \newtheorem{thmC}{Theorem}
  \newtheorem{corC}[thmC]{Corollary}
  \newtheorem{propC}[thmC]{Proposition}
\fi

\section{What a multi-view normal-map aggregate measures}
\label{sec:s_geometry}

The supervision of the main paper renders a shape from many views, applies a
2D detector to each rendered image, and aggregates the responses back onto the
surface. This section states what such an aggregate measures in an idealized
case, in which the per-view response is the trace of the structure tensor of the
rendered normal map. The results are taken from a companion analysis. We sketch
the proofs here, so that this document does not depend on that analysis.

\subsection{Setup}

Let $M\hookrightarrow\mathbb R^3$ be a smooth, compact, oriented surface with
unit normal field $N:M\to S^2$. At $q\in M$ the derivative of $N$ is the shape
operator $\dN_q:T_qM\to T_qM$. The shape operator is self-adjoint, and its
eigenvalues are the principal curvatures $\kappa_1\ge\kappa_2$. Hence
\begin{equation}
\|\dN_q\|_F^2=\kappa_1(q)^2+\kappa_2(q)^2=2\,C(q)^2,
\end{equation}
where $C(q)=\sqrt{(\kappa_1^2+\kappa_2^2)/2}$ is the curvedness defined by
Koenderink and van Doorn~\cite{koenderink1992surface}. The Gaussian curvature is
$K=\kappa_1\kappa_2$.

A viewing direction $\mathbf d\in S^2$ points from the surface toward the
camera. The normal map rendered from $\mathbf d$ stores at each pixel the unit
normal of the visible surface point. Its two image-plane channels are the
components of this normal orthogonal to $\mathbf d$. We write
$\mathbf S_{\mathbf d}(q)$ for the structure tensor of these two channels at the
pixel where $q$ is visible. For views $\mathbf d_1,\dots,\mathbf d_L$, let
$\mathcal V_L(q)$ be the set of indices $l$ for which $q$ is visible from
$\mathbf d_l$. The per-view response and the $L$-view aggregate are
\begin{equation}
H_{\mathbf d}(q)=\tr\bigl(\mathbf S_{\mathbf d}(q)\bigr),
\qquad
\rho_L(q)=\frac{1}{|\mathcal V_L(q)|}\sum_{l\in\mathcal V_L(q)}H_{\mathbf d_l}(q).
\label{eq:app_rho}
\end{equation}
As $L$ grows with uniform views, $\rho_L(q)$ converges to the mean of
$H_{\mathbf d}(q)$ over the directions from which $q$ is visible. The response
$H_{\mathbf d}$ does not change when $\mathbf d$ is replaced by $-\mathbf d$, and
a point on a convex surface is visible from exactly one hemisphere. On a convex
surface this mean therefore equals the mean over the whole sphere, which we
write $\mathbb E_{\mathbf d}[H_{\mathbf d}(q)]$. Corollary~\ref{cor:app_views} treats
other sets of visible directions.

The results below rest on four assumptions.

\begin{enumerate}
\item[(A1)] \emph{Orthographic projection.} The image is formed by orthographic
projection along $\mathbf d$.
\item[(A2)] \emph{Uniform views.} The viewing directions are distributed
uniformly on $S^2$.
\item[(A3)] \emph{Trace response.} The per-view response is the trace
$H_{\mathbf d}$ in~\eqref{eq:app_rho}. Section~\ref{sec:s_geo_det} treats the
determinant, which the classical Harris response uses.
\item[(A4)] \emph{Surface-length sampling.} Gradients are measured per unit
length on the surface. A pixel grid measures them per unit length in the image
instead. This enlarges the gradient along the foreshortened tangent direction by
the factor $1/|\langle N(q),\mathbf d\rangle|$, and the factor grows without
bound near the occluding contour.
\end{enumerate}

\subsection{The limit identity}

\begin{thmC}[Limit identity]
\label{thm:app_main}
Under \textup{(A1)--(A4)},
\begin{equation}
\mathbb E_{\mathbf d}\bigl[H_{\mathbf d}(q)\bigr]
=\tfrac23\,\|\dN_q\|_F^2
=\tfrac23\bigl(\kappa_1^2+\kappa_2^2\bigr)
=\tfrac43\,C(q)^2 .
\label{eq:app_main}
\end{equation}
\end{thmC}

\emph{Sketch.} Write $P_{\mathbf d}=I_3-\mathbf d\mathbf d^\top$. The two
image-plane channels store $P_{\mathbf d}N$. Under (A1) and (A4), a unit tangent
vector $v$ at $q$ changes these channels by $P_{\mathbf d}\,\dN_q(v)$. Let
$e_1,e_2$ be the principal directions at $q$, and let
$x=\langle e_1,\mathbf d\rangle$ and $y=\langle e_2,\mathbf d\rangle$. Then
\begin{equation}
\tr\mathbf S_{\mathbf d}
=\kappa_1^2\|P_{\mathbf d}e_1\|^2+\kappa_2^2\|P_{\mathbf d}e_2\|^2
=\kappa_1^2(1-x^2)+\kappa_2^2(1-y^2).
\label{eq:app_single}
\end{equation}
For a uniform direction on $S^2$, $\mathbb E[\mathbf d\mathbf d^\top]=\tfrac13I_3$,
so $\mathbb E[x^2]=\mathbb E[y^2]=\tfrac13$. The expectation is therefore
$\tfrac23(\kappa_1^2+\kappa_2^2)$. \hfill$\square$

By this identity, the aggregate ranks corners above edges and edges above flat
regions.
Curvedness does not depend on the signs of the principal curvatures, so the
aggregate does not distinguish a ridge from a rut.

\subsection{Non-uniform and occluded view sets}

In practice $q$ is seen only from part of the sphere, and the views are rarely
uniform on that part. Let
$\mathbf W(q)=\kappa_1^2\,e_1e_1^\top+\kappa_2^2\,e_2e_2^\top$. For every single
view, \eqref{eq:app_single} can then be written as
$H_{\mathbf d}(q)=\tr\mathbf W(q)-\mathbf d^\top\mathbf W(q)\,\mathbf d$.

\begin{corC}[Arbitrary view distributions]
\label{cor:app_views}
Under \textup{(A1)}, \textup{(A3)} and \textup{(A4)}, let $\mu_q$ be any
probability measure on the directions from which $q$ is visible, and let
$\Sigma_q=\mathbb E_{\mu_q}[\mathbf d\mathbf d^\top]$. Then
\begin{equation}
\mathbb E_{\mu_q}\bigl[H_{\mathbf d}(q)\bigr]
=\tr\mathbf W(q)-\tr\bigl(\Sigma_q\mathbf W(q)\bigr).
\label{eq:app_views}
\end{equation}
\end{corC}

\emph{Sketch.} Take the expectation of
$\tr\mathbf W(q)-\mathbf d^\top\mathbf W(q)\,\mathbf d$ under $\mu_q$.
\hfill$\square$

For the uniform distribution, $\Sigma_q=\tfrac13I_3$, and~\eqref{eq:app_views}
reduces to Theorem~\ref{thm:app_main}. When the views are drawn independently
from a distribution $\nu$ on $S^2$, $\rho_L(q)$ converges to the left side
of~\eqref{eq:app_views}, where $\mu_q$ is $\nu$ conditioned on the visibility of
$q$. The same formula covers cameras on a hemisphere, camera rings at
fixed elevations, and self-occlusion. The aggregate departs from
Theorem~\ref{thm:app_main} only through the second moment $\Sigma_q$ of the
directions from which $q$ was seen. This departure can therefore be computed
from a visibility record, without rendering the views again.

\subsection{How many views}

\begin{propC}[Finite-view estimate]
\label{prop:app_var}
Under \textup{(A1)--(A4)}, let $\rho_L(q)$ be the mean of $L$ responses
$H_{\mathbf d_l}(q)$ whose directions $\mathbf d_1,\dots,\mathbf d_L$ are drawn
independently and uniformly on $S^2$. Then $\rho_L(q)$ is unbiased for
$\tfrac23\|\dN_q\|_F^2$, and
\begin{equation}
\operatorname{Var}\bigl[\rho_L(q)\bigr]
=\frac{4}{45\,L}\Bigl(\|\dN_q\|_F^4-3K(q)^2\Bigr).
\label{eq:app_var}
\end{equation}
\end{propC}

\emph{Sketch.} The fourth moments of a uniform direction are
$\mathbb E[x^4]=\tfrac15$ and $\mathbb E[x^2y^2]=\tfrac1{15}$. With
\eqref{eq:app_single} they give
$\mathbb E[H_{\mathbf d}(q)^2]=\tfrac1{15}(8\|\dN_q\|_F^4-4K^2)$. Subtracting the
squared mean $\tfrac49\|\dN_q\|_F^4$ and dividing by $L$
gives~\eqref{eq:app_var}. \hfill$\square$

Dividing the single-view variance by the squared mean gives the relative
variance of a single view,
\begin{equation}
\frac{\operatorname{Var}[H_{\mathbf d}(q)]}{\mathbb E[H_{\mathbf d}(q)]^2}
=\frac15\Bigl(1-\frac{3K(q)^2}{\|\dN_q\|_F^4}\Bigr)\in\Bigl[\tfrac1{20},\tfrac15\Bigr].
\label{eq:app_relvar}
\end{equation}
The relative variance depends only on the ratio of the principal curvatures. It
is largest on edges, where $K=0$, and smallest where the two principal
curvatures have equal magnitude. The relative standard error of $\rho_L$ is
therefore at most $\sqrt{1/(5L)}$ at every point of every smooth surface. On a
smooth surface, the number of views needed for a given accuracy does not depend
on the shape.

Views at grazing angles are unreliable. If views with
$\langle N,\mathbf d\rangle<c_{\min}$ are discarded, the same moment computation
replaces the constant $\tfrac23$ in~\eqref{eq:app_main} by
$\tfrac12\bigl(1+\tfrac13(1+c_{\min}+c_{\min}^2)\bigr)$, which is $0.685$ at $c_{\min}=0.1$.
Under this cutoff, a numerical evaluation gives $0.174$ for the upper bound
in~\eqref{eq:app_relvar}. The relative standard error is then at most
$\sqrt{0.174/L}$, which is below $10\%$ for $L\ge18$.

Creases, thin shells and occlusion make the responses on CAD models vary more
from view to view than any smooth surface allows. On five
ShapeNet~\cite{chang2015shapenet} categories, the median single-view relative
variance over the half of the points with the larger mean response ranges from
$0.45$ to $0.98$. A relative error of $10\%$ then needs $45$ to $98$ visible
views. The bound in~\eqref{eq:app_relvar} is therefore only a lower limit on the
number of views for CAD models.

\subsection{The determinant and the classical Harris response}
\label{sec:s_geo_det}

\begin{propC}[Determinant-based response]
\label{prop:app_det}
Under \textup{(A1)}, \textup{(A2)} and \textup{(A4)},
$\det\mathbf S_{\mathbf d}(q)=K(q)^2\langle N(q),\mathbf d\rangle^2$ and
\begin{equation}
\mathbb E_{\mathbf d}\bigl[\det\mathbf S_{\mathbf d}-k(\tr\mathbf S_{\mathbf d})^2\bigr]
=\Bigl(\tfrac13+\tfrac{4k}{15}\Bigr)K^2-\tfrac{8k}{15}\|\dN_q\|_F^4 .
\end{equation}
\end{propC}

\emph{Sketch.} The projection from the tangent plane to the image plane scales
areas by $|\langle N(q),\mathbf d\rangle|$, so
$\det\mathbf S_{\mathbf d}(q)=K(q)^2\langle N(q),\mathbf d\rangle^2$. Under (A2),
$\mathbb E[\langle N(q),\mathbf d\rangle^2]=\tfrac13$. The second moment of
$H_{\mathbf d}(q)$ from the sketch of Proposition~\ref{prop:app_var} gives the
remaining term. \hfill$\square$

The expected determinant depends only on $K^2$, so it vanishes on developable
regions such as cylinders. The classical Harris response therefore ranks points
by a mixture of curvedness and Gaussian curvature, and its expectation is
negative along edges. Of the trace, the determinant and the classical Harris
response, only the trace has an aggregate proportional to the squared
curvedness.

\subsection{Deviations that matter in practice}

The identity is exact only under (A1)--(A4). The companion analysis measures
the departures on three analytic surfaces. The ellipsoid and the superquadric
are convex, and the torus hides part of its own surface in every view. The
departures below are listed in decreasing order of their effect on the
aggregate.

\begin{itemize}
\item Pixel-length gradients violate (A4) and have the largest effect. They
raise the median aggregate by a factor between $2.0$ and $2.7$, and the factor
varies widely across points. A view with $\langle N,\mathbf d\rangle=c$
shortens the surface by the factor $c$ in the image, so the squared gradient in
this direction grows by $1/c^2$. Only the grazing cutoff $c_{\min}$ limits this
factor. The rank correlation with curvedness stays at $0.99$ on the convex
surfaces and drops to $0.44$ on the torus.
\item Perspective projection violates (A1). At camera distances of about $7$ and
$3$ times the object radius, it raises the median aggregate by a factor between
$1.05$ and $1.24$. The excess decays roughly as $1/d$ with the camera
distance~$d$. The ranking on the convex surfaces is unchanged. On the torus, the
largest value of $\kappa_1^2+\kappa_2^2$ is only $1.36$ times the smallest. The
excess also varies over the torus, and this variation reorders the points.
\item Non-uniform views violate (A2). With cameras on the upper hemisphere, on
three rings or on a single ring, the median ratio of the aggregate to the
prediction of Corollary~\ref{cor:app_views} lies between $1.001$ and $1.006$.
The rank correlation stays above $0.92$ on the convex surfaces. On the torus it
stays at $0.94$ with three rings and drops to about $0.4$ with the upper
hemisphere or a single ring.
\item Discretization of the image leaves a small bias. With the cutoff
$c_{\min}=0.1$, the median ratio of the aggregate to $\kappa_1^2+\kappa_2^2$ lies
between $0.685$ and $0.690$ at $1024$\,px, against the predicted $0.685$. This
ratio approaches $0.685$ as the resolution grows, and it changes by less than
$0.01$ between $25$ and $400$ views.
\end{itemize}

A learned 2D detector falls outside the theorem. The companion analysis also
aggregates the detection scores of SuperPoint~\cite{detone2018superpoint} over
the same views and visibility masks as the trace response. This aggregate
correlates positively with curvedness on all three surfaces, but less than the
trace response does. The rank correlations are $0.77$, $0.87$ and $0.34$, and
the trace response in the same run reaches $1.00$, $1.00$ and $0.92$.
Multi-view supervision from such a detector therefore follows curvedness only
in part. Section~III-E of the main paper relies on this statement.

\section{Additional results}
\label{sec:s_results}

\subsection{Real scenes: all evaluation crops}
\label{sec:s_sf3d_all}

Table~\ref{tab:s_sf3d_all} repeats the localization success of Table~III of the
main paper on all 30 evaluation crops. These crops include the six crops used
during development and give 2640 frame pairs per budget. The margins of
\ours{} over the other methods remain significant in every column.

\begin{table}[h]
\centering
\caption{Real indoor scenes: localization success (\%) on all evaluation crops}
\label{tab:s_sf3d_all}
\footnotesize
\begin{threeparttable}
\setlength{\tabcolsep}{4pt}
\begin{tabular*}{\columnwidth}{@{\extracolsep{\fill}}lccc@{}}
\toprule
Method & $5^\circ$,\,5\,cm & $5^\circ$,\,10\,cm & $10^\circ$,\,25\,cm\\
\midrule
\textbf{\ours{}} & \textbf{36.5} & \textbf{57.5} & \textbf{72.9}\\
Harris-3D~\cite{sipiran2011harris} & 30.7 & 51.3 & 68.5\\
D3Feat~\cite{bai2020d3feat} & 9.0 & 21.3 & 37.8\\
ISS~\cite{zhong2009iss} & 8.3 & 20.4 & 38.4\\
SC3K~\cite{zohaib2023sc3k} & 8.2 & 20.3 & 37.9\\
USIP~\cite{li2019usip} & 7.7 & 19.9 & 37.4\\
Key-Grid~\cite{hou2024keygrid} & 6.4 & 17.2 & 33.9\\
UKPGAN~\cite{you2022ukpgan} & 4.1 & 12.5 & 28.4\\
Random & 6.8 & 17.0 & 34.0\\
\bottomrule
\end{tabular*}
\begin{tablenotes}[para,flushleft]\footnotesize
\item[]\hspace{-\labelsep}The evaluation uses all 30 SceneFun3D~\cite{delitzas2024scenefun3d}
evaluation crops, including the six crops used during development, and 2640 real
frame pairs. Success is averaged over $K\in\{64,80,96\}$.
\end{tablenotes}
\end{threeparttable}
\end{table}

\subsection{Sparse objects: all spacings}
\label{sec:s_sparse_objects}

Table~\ref{tab:s_sparse_objects} lists the values plotted in Fig.~8 of the main
paper. The parameters of the baselines are chosen separately for each spacing on
these test objects. From the smallest to the largest spacing, Harris-3D uses
$k=16$, $16$, $8$, $8$, $8$ and $32$ nearest neighbours. Continuous Harris-3D
uses $k=16$, $16$, $8$, $8$, $8$ and $8$. ISS uses a radius of $1.5\,s$ at every
spacing.

\begin{table}[h]
\centering
\caption{Sparse objects: localization success (\%)}
\label{tab:s_sparse_objects}
\footnotesize
\begin{threeparttable}
\setlength{\tabcolsep}{3.5pt}
\begin{tabular*}{\columnwidth}{@{\extracolsep{\fill}}lcccccc@{}}
\toprule
Method & $s{=}0.01$ & $0.02$ & $0.03$ & $0.04$ & $0.05$ & $0.06$\\
\midrule
\textbf{\ours{}} & \textbf{34.6} & \textbf{33.8} & \textbf{26.4} & \textbf{24.7} & \textbf{16.9} & \textbf{16.2}\\
Harris-3D & 21.5 & 16.0 & 9.9 & 9.0 & 6.5 & 6.6\\
Continuous Harris-3D & 21.7 & 15.8 & 10.1 & 9.1 & 6.4 & 6.4\\
ISS & 16.1 & 12.2 & 7.8 & 7.4 & 5.0 & 5.2\\
Random & 4.4 & 4.8 & 4.1 & 5.0 & 4.3 & 5.0\\
\bottomrule
\end{tabular*}
\begin{tablenotes}[para,flushleft]\footnotesize
\item[]\hspace{-\labelsep}The evaluation uses 130 held-out ShapeNet models from 15 categories,
thinned to point spacing $s$. The longest side of each model is $0.9$ in normalized coordinates.
Success is averaged over $K\in\{32,64,128,256\}$, and a pose counts as a success
within $10^\circ$ and $0.05$.
\end{tablenotes}
\end{threeparttable}

\end{table}

\subsection{Tuning Harris-3D}
\label{sec:s_tune}
The only free parameter of Harris-3D is the neighbourhood over which the quadric
is fitted. The standard $k=32$ need not be the best choice at our scale, so we
tuned this neighbourhood on the test sets. This tuning gives the baseline an
advantage that we do not take for \ours{}.

\paragraph{Objects}
We re-ran Harris-3D with $k=128$ and $512$ nearest neighbours and with fixed
radii of $0.02$, $0.04$ and $0.08$ in normalized coordinates. For each budget, we
kept the setting that scored highest on the test objects. On localization
success, the standard $k=32$ remains the best setting at every budget. On hit
rate, the $0.02$ radius is the best setting. The margins of \ours{} over this
setting are $+12.3$, $+14.6$ and $+16.5$ points at $K=32$, $16$ and $8$, and
every interval excludes zero.

\paragraph{Real scenes}
We compared $k=16$, $32$, $64$ and $128$ nearest neighbours and fixed radii of
$2$, $4$ and $8$\,cm. The $2$\,cm radius is the best setting, with a success of
$55.2\%$ under the primary thresholds. The margins of \ours{} over this setting
are $+3.4$, $+4.5$ and $+3.9$ points under the strict, primary and loose
thresholds. All three intervals exclude zero, and the strict one only just.

\subsection{Further ablations}
\label{sec:s_ablation}

\paragraph{Test-time averaging}
On real scenes, the field is averaged at test time over the input and three
random rotations of it, each with two random encoder subsamples. Table~VI of the
main paper evaluates this averaging on the full model trained for $80$ epochs.
The averaging adds $1.0$ point under the primary thresholds, and the interval
contains zero. Without the averaging, the model still outperforms Harris-3D by
$4.9$ points, and this interval excludes zero.

\paragraph{The same features on objects}
On objects, the per-query features also help. We compare with a model trained
for the same number of epochs without them. The features add $2.8$ and $3.8$
points of hit rate at $K=32$ and $16$, and $2.3$, $3.9$ and $2.3$ points of
localization success at $K=256$, $128$ and $64$. All these differences are
significant.

\subsection{Pose error on real scenes}
\label{sec:s_err}
Section~IV-C of the main paper compares the geometric mean of the pose error of
\ours{} and Harris-3D over all frame pairs at $K=80$. This comparison uses the
pose error of every frame pair, including the poses outside the success
thresholds. When too few matches remain, PnP returns no pose, and the pair has
no pose error. This happens for $265$ of the $2100$ pairs with \ours{} and for
$339$ with Harris-3D. We assign these pairs an error of $10$\,m and
$180^\circ$. Errors below $1$\,mm and $0.01^\circ$ are raised to these values
before the logarithm. With these settings, the geometric mean of the
translation error of \ours{} is $0.76\times$ that of Harris-3D, and the
geometric mean of the rotation error is $0.75\times$. With a milder value of
$1$\,m and $30^\circ$ for the missing poses, the ratios are $0.82\times$ and
$0.80\times$. All four ratios are significant under the bootstrap over crops.

\section{Locating corners between samples}
\label{sec:s_corners}

\begin{figure*}[t]
\centering
\includegraphics[width=\textwidth]{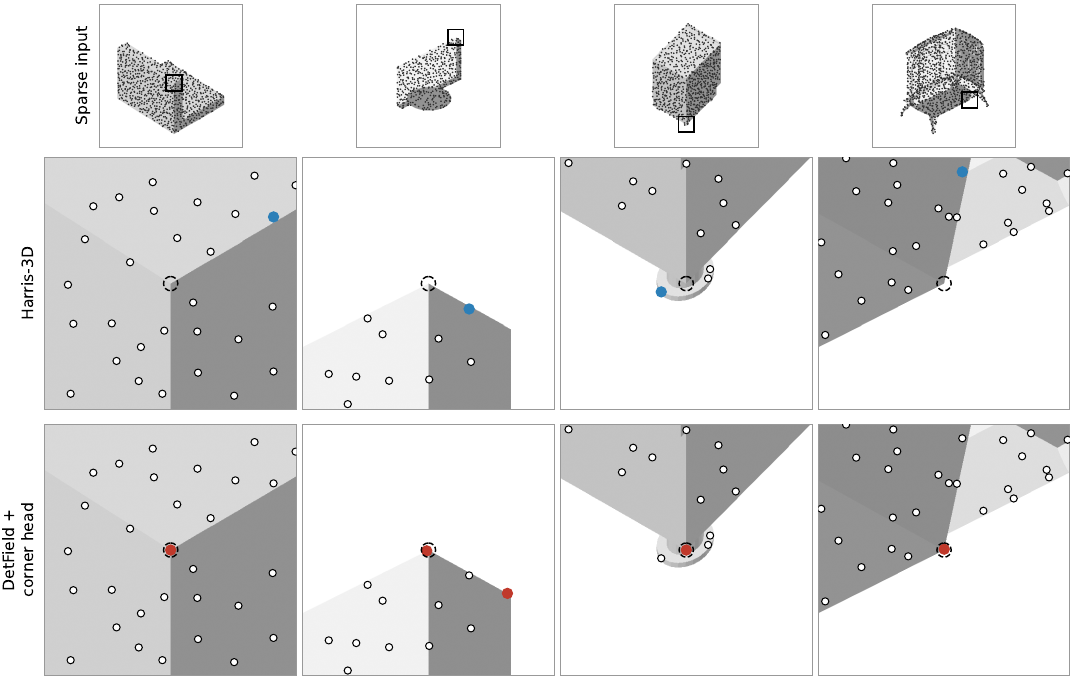}
\caption{\label{fig:s_offsample}Locating corners between the samples of four
sparse held-out objects at $s=0.04$. Top: the sparse input, with the zoomed
region marked. Middle and bottom: the zoomed region with the sparse points
(white) and the CAD corner (dashed circle). Coloured points are the points
selected by Harris-3D (middle) and by \ours{} with the corner head (bottom).
Harris-3D can only select among the samples, while the corner head places a
point on the corner. The examples are chosen among the objects in which the
nearest sample lies at least $s/3$ from the corner.}
\end{figure*}

A continuous representation can also locate a landmark that no sample lies on.
We test this on sparse objects with a corner head added to \ours{}. At any query
location, the corner head predicts an offset towards the nearest corner. These
predictions are combined with local plane fits among the sparse points, and the
resulting candidates are ranked. Fig.~\ref{fig:s_offsample} shows examples at
$s=0.04$, in which the nearest sample lies at least $s/3$ from the true corner.
Harris-3D and every other method that selects among the samples therefore miss
the corner under the $s/4$ criterion of Table~\ref{tab:s_corners}. The corner
head places a point on the corner. The localization experiments of the main
paper select among the sampled points, so this experiment evaluates the position
of a landmark on its own.

\paragraph{Corner head}
At any query location, a second head on the network of the main paper predicts
the confidence that a corner lies within $R=0.08$ and the offset towards this
corner. The network is initialized from the object model of the main paper and
trained in two stages, with $1000$ samples per epoch. The first stage uses
synthetic shapes assembled from boxes and plates. These shapes are tables,
chairs and groups of two to six boxes, and their convex corners are known
exactly. This stage runs for $40$ epochs at spacings drawn from
$[0.025, 0.06]$. The second stage fine-tunes the network for $20$ epochs at
spacings drawn from $[0.02, 0.05]$ on $1858$ further ShapeNet models, and $30\%$
of its samples are still synthetic. These $1858$ models come from $16$
categories and are disjoint from the $805$ models of the main paper and from the
$130$ test models. On ShapeNet, the ground-truth corners are the visible mesh
vertices where at least three large planar regions meet. The normals of three of
these regions must form a matrix whose determinant exceeds $0.3$ in magnitude.

\paragraph{Candidates and ranking}
The candidates come from three sources. The first source fits three planes by
local RANSAC~\cite{fischler1981random} in a neighbourhood of each sparse point
and takes their intersection. It uses neighbourhood radii of $2.5\,s$, $3\,s$
and $4\,s$. An intersection is kept when the three normals form a matrix whose
determinant exceeds $0.3$ in magnitude and the intersection lies within
$1.5\,s$ of the sparse point. The second source runs the same plane fit within
$s$ of each point that the corner head votes for. The third source is the voted
points themselves. A gradient-boosted classifier scores each candidate by
whether it lies within $s/4$ of a corner. Its features include the vote support,
the confidence and offset predicted by the corner head, the plane-fit
statistics, the distance to the nearest sample, and the number of neighbourhood
radii that produce an intersection at the same place. The classifier is trained
on $289$ of the $1858$ fine-tuning models. Neither the corner head nor the
classifier uses the test models. The candidates are selected in the order of
their scores, with non-maximum suppression of radius $s/2$.

\paragraph{Evaluation}
Each method selects as many points as the object has ground-truth corners. The
selected points are matched one to one to the corners with the minimum total
distance. Table~\ref{tab:s_corners} reports the fraction of corners matched
within $s/4$. \emph{Best on samples} assigns to each corner its nearest sparse
point, so it bounds any method that selects among the samples.

\begin{table}[h]
\centering
\caption{Corner head: corners located within $s/4$ (\%)}
\label{tab:s_corners}
\footnotesize
\begin{threeparttable}
\begin{tabular*}{\columnwidth}{@{\extracolsep{\fill}}lcccc@{}}
\toprule
Method & $s{=}0.02$ & $0.03$ & $0.04$ & $0.05$\\
\midrule
\textbf{Corner head} & \textbf{40.1} & \textbf{34.1} & \textbf{33.5} & \textbf{29.0}\\
Best on samples (bound) & 18.4 & 16.1 & 21.7 & 16.9\\
Local three-plane RANSAC & 25.7 & 21.1 & 21.7 & 20.2\\
Harris-3D ($k{=}16$) & 1.6 & 1.9 & 3.4 & 1.9\\
ISS ($1.5\,s$) & 0.2 & 0.5 & 0.3 & 0.5\\
\bottomrule
\end{tabular*}
\begin{tablenotes}[para,flushleft]\footnotesize
\item[]\hspace{-\labelsep}The evaluation uses 130 held-out ShapeNet models with 1726 corners in
total. Local three-plane RANSAC uses the first candidate source with the single
radius $3\,s$ and ranks the candidates by their smallest inlier count. Harris-3D
and ISS select among the samples with non-maximum suppression of radius $s/2$.
\end{tablenotes}
\end{threeparttable}
\end{table}

\section{Baseline sanity}
\label{sec:s_sanity}

In the main tables, the points selected by the five learned 3D keypoint methods
are re-detected in images at about the same rate as random points. A possible
explanation is that these methods were trained badly. We checked them in two
ways. First, we checked the training curve and the output of each method.
Second, we tested the trained USIP model on its own metric.

\paragraph{Convergence checks}
USIP, D3Feat, SC3K and Key-Grid are trained on our data. On objects, the
validation loss of USIP decreases from $-4.40$ to $-5.88$ over $200$ epochs. The
validation descriptor loss of D3Feat decreases from $5.82$ to $3.73$ over $80$
epochs, and its weights load with no missing or unexpected parameters. The
validation loss of SC3K decreases from $87$ to $32$ over $60$ epochs, and its
keypoints spread over the object instead of collapsing to a single point. The
validation loss of Key-Grid decreases from $0.188$ to $0.070$ over $25$ epochs.
On real-scene crops, the validation descriptor loss of D3Feat decreases from
$4.18$ to $3.46$ over $60$ epochs, and USIP reaches a validation loss of $-5.33$.
The keypoints of SC3K and Key-Grid again spread over the crop instead of
collapsing to a single point. Key-Grid places several keypoints at the same
position, with $18$ distinct positions out of $32$ on objects and $61$ out of
$128$ on crops.
UKPGAN uses the weights released by its authors.
On airplanes, its saliency concentrates on the engines, the tail and the wing
roots. These checks show that each pipeline runs as intended. They do not show
how good the trained models are.

\paragraph{Repeatability of USIP under rotation}
We use the USIP weights of the main tables. They are trained with full $SO(3)$
rotation augmentation, as in the original ModelNet~\cite{wu20153d} setting of
USIP. Each of the 60 evaluation objects is rotated by three random rotations
about its centre. The same points are rotated without resampling, as in the
rotation test of USIP. On the original and on each rotated copy, every method
selects $K$ keypoints with the non-maximum suppression of radius $0.02$ used in
the main paper. The relative repeatability is the fraction of the $K$ keypoints
of the original pose that have a keypoint of the rotated copy within
$\varepsilon$ after the rotation is undone. Table~\ref{tab:s_usip} reports the
values at $\varepsilon=0.02$.

At $K=32$, USIP exceeds ISS by $49.7$ points and random points by $60.3$ points.
Both $95\%$ intervals exclude zero under the bootstrap over objects. The other
budgets and the tolerances $\varepsilon=0.01$ and $0.04$ give the same ordering.
USIP has therefore learned keypoints that are stable under rotation, and its
training on our data is effective.

Harris-3D is a deterministic and exactly rotation-invariant function of the
point neighbourhood. On a rotated copy of the same points, it selects the same
points. Its repeatability under this protocol is therefore close to $100\%$ by
construction and carries no information. ISS is low partly for an
implementation reason. The Open3D~\cite{zhou2018open3d} implementation of ISS
returns candidates without scores, so we take $K$ of them by farthest-point
sampling. The order of the candidates changes with the rotation, and the
starting point of the sampling changes with it.

\begin{table}[h]
\centering
\caption{Baseline sanity: relative repeatability (\%) under rotation}
\label{tab:s_usip}
\footnotesize
\begin{threeparttable}
\begin{tabular*}{\columnwidth}{@{\extracolsep{\fill}}lccc@{}}
\toprule
Method & $K{=}16$ & $32$ & $64$\\
\midrule
USIP & 64.8 & 67.1 & 74.6\\
ISS & 17.2 & 17.4 & 20.6\\
Random & 3.8 & 6.8 & 11.6\\
\midrule
Harris-3D$^\dagger$ & 99.9 & 100.0 & 100.0\\
\bottomrule
\end{tabular*}
\begin{tablenotes}[para,flushleft]\footnotesize
\item[]\hspace{-\labelsep}The evaluation uses the 60 evaluation objects with three random rotations
each. The tolerance is $\varepsilon=0.02$ in normalized coordinates.
\item[$\dagger$] Trivial under this protocol (see text).
\end{tablenotes}
\end{threeparttable}
\end{table}

\paragraph{Deviations from the original settings}
USIP, D3Feat, SC3K and Key-Grid are trained on our objects. For the real-scene
experiments, they are trained again on our training crops. The original D3Feat
is trained on 3DMatch~\cite{zeng20173dmatch}. The original SC3K and Key-Grid are
trained separately for each category, and here each of them is trained as a
single model across categories. Both methods also tie the number of keypoints to
the last layer, so they appear only at budgets up to the number they were
trained for. This number is $32$ on objects and $128$ on real scenes. UKPGAN uses
the universal weights released by its authors, without retraining on our data.

\section{Implementation details}
\label{sec:s_impl}

This section lists the constants of the pipeline. Section~III of the main paper
gives only the ones that the method depends on. Unless a physical unit is given,
coordinates are in normalized units, in which the longest side of an object or
crop is $0.9$.

\subsection{Network}
\label{sec:s_impl_net}

\paragraph{Architecture}
The feature volumes at $32^3$, $64^3$ and $128^3$ have $128$, $64$ and $32$
channels, and each U-Net~\cite{cciccek20163dunet} has three levels. The hidden
width is $256$. The encoder and the decoder both have five residual blocks, and
the feature volumes use a padding of $0.1$. The model has $6.6$\,M parameters.
The input point cloud is normalized so that $\max_i\|x_i\|_\infty=0.45$. During
training, the encoder sees a random subsample of $4096$ points for objects and
$8192$ points for scene crops. At test time, these numbers are $8192$ and
$20\,000$. On dense maps, the decoder is evaluated at every input point.

\paragraph{Curvature features}
For each point, the $64$ nearest neighbours are found on the full-resolution
cloud. The scale $k\in\{16,32,64\}$ uses the first $k$ of them, including the
point itself. The neighbourhood is centred at its mean and expressed in its
principal frame. A quadric $z=a_0+a_1u+a_2v+a_3u^2+a_4uv+a_5v^2$ is fitted by
least squares, which gives $f_{uu}=2a_3$, $f_{uv}=a_4$ and $f_{vv}=2a_5$. The
features are $f_{uu}^2+2f_{uv}^2+f_{vv}^2$, $(f_{uu}f_{vv}-f_{uv}^2)^2$ and the
ratio $\lambda_{\min}/\sum\lambda$ of the neighbourhood covariance, each at the
three values of $k$. Each channel is replaced by its rank within the sample
divided by the number of points. This normalization is computed on the full
sample before any subsampling. A query point takes the features of its nearest
encoder point. On sparse maps, the features are recomputed on the sparse points.

\paragraph{Input sampling}
Points are sampled over the whole repaired mesh, including interior structure.
Some of these points are structurally invisible, because none of the $128$
views sees them. Over the $799$ training and evaluation objects, they make up
$6.6\%$ of the points on average and $0.8\%$ in the median shape, and the worst
shape reaches $60\%$. Microwaves and cars are affected most, with medians of
$17\%$ and $10\%$. The point density on the surface is fixed, and the encoder
subsamples to a fixed number of points. A shape with $10^5$ points therefore
receives about three times fewer encoder points per unit area than a shape with
$3\times10^4$ points.

\subsection{Supervision on objects}
\label{sec:s_impl_obj}

\paragraph{Rendering and baking}
Table~\ref{tab:s_bake} lists the settings of the supervision pipeline of
Section~III-C. The detector map is the per-cell softmax of SuperPoint with the
dustbin removed. It is taken before any threshold or non-maximum suppression,
and it is computed on the encoded image, which has a single grey channel. The
depth-spread term $\rho$ of the occlusion measure is the range of depths in the
$9\times9$ window minus $9\,\|\nabla d\|$, where $\nabla d$ is the
central-difference depth gradient.

\begin{table}[h]
\centering
\caption{Supervision: rendering and baking settings}
\label{tab:s_bake}
\footnotesize
\begin{tabular}{@{}>{\raggedright\arraybackslash}p{0.27\columnwidth}>{\raggedright\arraybackslash}p{0.67\columnwidth}@{}}
\toprule
Normalization & bounding-box centre at origin, longest side $0.9$\\
Normals & deduplicated, consistently oriented, crease $30^\circ$\\
Surface points & objects $6\times10^4$ per unit area ($2000$ to $3\times10^5$);
crops $4.8\times10^5$ per unit area (at most $7\times10^5$)\\
Views & $128$, Fibonacci lattice; vertical field of view $60^\circ$, aspect $4{:}3$\\
Framing & bisection per view until the projected extent fills $0.88$ of the image\\
Resolution & objects $480\times640$, $3\times$ supersampling; crops $960\times1280$, none\\
Encoding & $0.7\,n_z+0.3\,d_{01}$, background $0$\\
Mirror averaging & $\tfrac12(\mathrm{SP}(I)+F[\mathrm{SP}(F[I])])$\\
Occlusion measure & $9\times9$ window; $b$: $0.15$, $0.08$; $\rho$: $0.02$, $0.01$\\
Reweighting & $\lambda=2$, clamped to $[0,1]$\\
Reading & maximum over a $9\times9$ window at the rounded projection\\
Visibility & $|z-z_{\mathrm{render}}|<0.02$\\
Depth-span filter & drop an observation whose $9\times9$ foreground depth span exceeds $0.10$\\
Aggregation & mean over the retained observations\\
Reliability weight & $\min(n/c_0,1)$, $c_0=6$, $n$ = visible views before the filter\\
\bottomrule
\end{tabular}
\end{table}

\begin{figure}[h]
\centering
\includegraphics[width=\columnwidth]{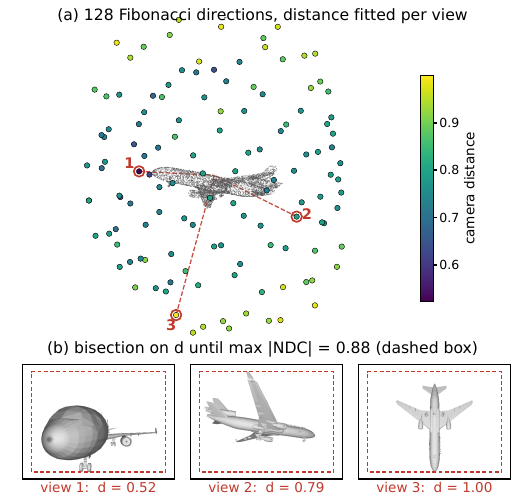}
\caption{\label{fig:s_views}View configuration of the supervision. (a) The $128$
cameras lie on a Fibonacci lattice around the object. The distance of each camera
is set independently by bisection, so it varies with direction (colour). The camera
centres therefore do not lie on a sphere. Cameras above and below the airplane
are farthest, because the object shows its largest extent in these directions. (b) For the
nearest, median and farthest view, the projected object reaches $0.88$ of the
image half-extent (dashed box). The views are rendered from the mesh. In view 1,
the camera lies $0.07$ in front of the nose, so perspective makes the nose appear
much larger than the rest of the airplane.}
\end{figure}

\paragraph{View configuration}
Figure~\ref{fig:s_views} shows the camera layout of Section~III-C. Object images
are rendered at $480\times640$ with $3\times$ supersampling.

\paragraph{Per-view corrections}
The detector is not equivariant to mirroring. On our renderings, the difference
between $\mathrm{SP}(I)$ and the mirrored response of the mirrored image reaches
$70\%$ of the mean response. The symmetrized response is
\begin{equation}
r = \tfrac12\bigl(\mathrm{SP}(I)+\mathcal F[\mathrm{SP}(\mathcal F[I])]\bigr),
\end{equation}
where $\mathcal F$ is the horizontal flip. The local occlusion measure is
\begin{equation}
\hat\delta=\max\Bigl(\mathrm{sig}\bigl(\tfrac{b-0.15}{0.08}\bigr),\;
                     \mathrm{sig}\bigl(\tfrac{\rho-0.02}{0.01}\bigr)\Bigr),
\end{equation}
where $b$ is the fraction of background pixels in a $9\times9$ window and $\rho$
is the depth spread in that window after subtracting the spread that the local
depth gradient alone would produce. Without that subtraction, a slanted but
continuous surface, such as a wing seen edge-on, is indistinguishable from an
occlusion boundary. The reweighted response is
\begin{equation}
r' = \mathrm{clamp}\Bigl(r\cdot\exp\bigl(\lambda\,[2\hat\delta-1]_+(1-r)
     -\lambda_{\mathrm{s}}\,[1-2\hat\delta]_+\bigr),\,0,\,1\Bigr),
\label{eq:s_dw}
\end{equation}
with $\lambda=\lambda_{\mathrm{s}}=2$. The factor $(1-r)$ keeps already strong
responses from being amplified further. The clamp keeps $r'$ in the range of
the quantity that it will become.

Observations whose local depth spread exceeds $0.10$ are discarded. In a probe
on six models, this filter removed about $9.5\%$ of the observations and raised
the consistency of the aggregate across views. Discarding the same fraction of
observations at random gave a smaller gain, so the discarded observations are
less reliable than average. The threshold of $0.10$ gave the highest
consistency, and tighter thresholds began to remove observations at real
geometric edges.

\paragraph{Reliability weight}
The weight of Eq.~(6) of the main paper divides the number of views in which a
point is visible by $c_0=6$ and caps the result at $1$. The constant $c_0$ is
chosen by hand. For the trace response, Section~\ref{sec:s_geometry} gives the
variance of a finite-view aggregate in closed form. A weight derived from this
variance could replace the cap.

\subsection{Supervision on real scenes}
\label{sec:s_impl_scene}

Crops are cut from the SceneFun3D laser scan, which has a point spacing of about
$5$\,mm. Along the camera trajectory, a frame becomes a seed when the camera has
moved at least $0.5$\,m since the previous seed. The cluster of a seed collects
the frames within $15^\circ$ and $0.5$\,m of it. The cluster is thinned to at
most five frames around the seed, and it must contain at least one pair of frames
that are $5$--$10^\circ$ and $0.15$--$1.0$\,m apart. A cluster is kept when the
laser points inside its frusta, up to the $99$th percentile of the frame depths,
number at least $20\,000$, span a longest side of $2$ to $6.5$\,m, and satisfy
$\lambda_3/\sum\lambda\ge0.03$.

The laser scan of each room is reconstructed once by Poisson
reconstruction~\cite{kazhdan2013screened} at depth $11$, and the vertices with
the lowest $3\%$ of the density are removed. The crop mesh is the part of this
mesh inside the frusta, extended by a margin of $0.3$\,m and decimated to at most
$3\times10^5$ faces. The point cloud of the crop is sampled from this mesh, as
listed in Table~\ref{tab:s_bake}. Responses near an open boundary of the crop
mesh are attenuated by
\begin{equation}
w=\min\bigl(e^{-4\,[1-d_{\mathrm{out}}/0.117]_+},\;e^{-2\,[1-d_{\mathrm{hole}}/0.06]_+}\bigr),
\end{equation}
where $d_{\mathrm{out}}$ and $d_{\mathrm{hole}}$ are the geodesic distances in
metres to the outer boundary and to the nearest hole. The attenuation therefore
reaches $11.7$\,cm from the outer boundary and $6$\,cm from a hole.

The photographic target is baked from $60$ frames of the video of the room at
half resolution. These are the frames that see most of the crop, after the
evaluation frames and all frames within $0.3$\,m of them are excluded. A point
counts as observed in a frame when its depth agrees with the depth map of the
frame within $5$\,cm, and the response is read with the same $9\times9$ maximum.
This target has no mirror averaging, occlusion reweighting or depth-span filter.
The training target ranks the observed points by the photographic response and
the unobserved points by the rendered target, each within its own subset. These
ranks are then mapped onto the sorted values of the rendered target.

\subsection{Training}
\label{sec:s_impl_train}

\begin{table}[h]
\centering
\caption{Training settings}
\label{tab:s_train}
\footnotesize
\begin{threeparttable}
\setlength{\tabcolsep}{3pt}
\begin{tabular}{@{}>{\raggedright\arraybackslash}p{0.24\columnwidth}>{\raggedright\arraybackslash}p{0.2\columnwidth}>{\raggedright\arraybackslash}p{0.2\columnwidth}>{\raggedright\arraybackslash}p{0.27\columnwidth}@{}}
\toprule
& Objects (1) & Objects (2) & Scenes\\
\midrule
Initialization & random & stage 1 & stage 1\\
Epochs & $30$ of $100$ & $30$ & $160$\\
Batch / points & $2$ / $4096$ & $2$ / $4096$ & $4$ / $8192$\\
Features & none & $9$, per query & $9$, per query\\
Augmentation & none & none & rotation, translation $0.05$, noise $0.005$, keep $50$--$100\%$\\
\bottomrule
\end{tabular}
\begin{tablenotes}[para,flushleft]\footnotesize
\item[]\hspace{-\labelsep}All models use AdamW~\cite{loshchilov2019adamw} with zero weight decay, a learning rate of $10^{-4}$ with
cosine decay to $10^{-6}$, seed $42$ and $8192$ queries. Objects (1) and (2) are
the two training stages.
\end{tablenotes}
\end{threeparttable}
\end{table}

Table~\ref{tab:s_train} lists the training settings of the three models of the
main paper. Of the queries, $75\%$ lie near the surface, and $25\%$ are drawn
uniformly in the cube $[-0.5,0.5]^3$. A near-surface query starts from a base
point drawn uniformly over the occupied $0.03$ voxels of the input. The base
point is then offset by $\mathcal N(0,(0.01\,u^2)^2)$ with
$u\sim\mathcal U(0,1)$. Queries farther than $2\sigma$ from the input are
masked, with $\sigma=0.01$.

The sparse-map model starts from the scene model and is trained for $90$
epochs. Each step processes one crop, and gradients are accumulated over four
steps. The point spacing is drawn from $[10,120]$\,mm, the encoder sees at most
$20\,000$ points, and each crop receives a uniform rotation and a translation of
up to $0.05$. Of its queries, $40\%$ are dense points and $40\%$ are candidate
positions, namely six per sparse point plus the sparse points themselves. The
remaining $20\%$ are dense points perturbed by $\mathcal N(0,(2\sigma)^2)$ with
$\sigma=1$\,cm. A query farther than $3\sigma$ from the dense scan is a negative
with weight one.

\subsection{Inference and localization}
\label{sec:s_impl_loc}

Objects use a single pass with $8192$ encoder points. On scenes, the within-crop
rank of the field is averaged over the input and three random rotations of it
about its centroid. Each of these four copies uses two encoder subsamples of
$20\,000$ points. Sparse maps use the same four rotations and encode all sparse
points. At test time, each sparse point contributes $12$ candidate positions in
addition to itself. The candidates are drawn uniformly in the tangent disc of
radius $0.75\,s$ and offset along the normal by $\mathcal U(-0.25\,s,0.25\,s)$.
The normal is estimated from the $16$ nearest sparse points. Keypoints are
extracted by greedy radius suppression (Section~III-G). The hit-rate table on
objects uses small budgets, and its suppression radius is $0.05$ instead of
$0.02$ for all methods. Table~\ref{tab:s_loc} lists the settings of the localization pipeline.

Query views on objects are constructed at exact angular offsets from the
reference view. A spherical lattice of $L$ directions cannot produce pairs closer
than about $\sqrt{4\pi/L}$, and the informative range of offsets is below this
spacing. Descriptors are read at the projected location of each 3D point without
snapping to the nearest 2D detection. Snapping discards the points that fail to
snap and makes every method worse. On real scenes, the keypoints are selected
among the points visible in the reference frame, so that every method enters PnP
with the same number of candidates.
The hit tolerance is $8$\,px on real scenes and $4$\,px on objects. On real
scenes, a 3D offset of $2$\,cm at a distance of $2$\,m projects to about $8$\,px
in the half-resolution photographs. The larger tolerance on real scenes covers
residual errors of this size in the reconstructed geometry and in the camera
poses.

\begin{table}[h]
\centering
\caption{Localization pipeline settings}
\label{tab:s_loc}
\footnotesize
\setlength{\tabcolsep}{3pt}
\begin{tabular}{@{}>{\raggedright\arraybackslash}p{0.22\columnwidth}>{\raggedright\arraybackslash}p{0.34\columnwidth}>{\raggedright\arraybackslash}p{0.34\columnwidth}@{}}
\toprule
& Objects & Real scenes\\
\midrule
2D detections & NMS $4$\,px, threshold $0.005$, top $1024$ & NMS $4$\,px, threshold $0.002$, top $4096$\\
Descriptor & \multicolumn{2}{>{\raggedright\arraybackslash}p{0.72\columnwidth}}{SuperPoint, bilinear at the projected 3D point, no snapping}\\
Matching & \multicolumn{2}{>{\raggedright\arraybackslash}p{0.72\columnwidth}}{nearest neighbour, Lowe ratio~\cite{lowe2004distinctive} $0.9$, no mutual check}\\
PnP & EPnP~\cite{lepetit2009epnp} RANSAC, $2000$ iterations, $4$\,px & EPnP RANSAC, $2000$ iterations, $8$\,px\\
Images & $480\times640$ renders & photographs at half resolution\\
Pairs & $12$ base views $\times$ offsets $\{5,10,20,30\}^\circ$ ($\{5,10,20\}^\circ$ for hit rate) & up to $10$ frames per crop, all ordered pairs\\
Visibility & depth within $0.03$ & depth within $0.05$\,m\\
Hit tolerance & $4$\,px & $8$\,px\\
Success & $10^\circ$, $0.05$ & $5^\circ$, $10$\,cm; also $5^\circ$, $5$\,cm and $10^\circ$, $25$\,cm\\
\bottomrule
\end{tabular}
\end{table}

Inference was timed on one RTX 5090 with no other process on the GPU. On the
$24$ test crops, which hold $4.2\times10^5$ points in the median, one pass of
the field takes $0.06$\,s. The eight passes over rotations and subsamples take
$0.8$\,s, and $1.4$\,s on the largest crop. Peak GPU memory is $1.5$\,GB. On
objects, one pass takes $0.03$\,s. These times do not include the nine
geometric features, which are computed once per input beforehand. Training on one RTX 5090 takes about
$20$\,min and $53$\,min for the two object stages, $2.4$\,h for the scene model
and $2.4$\,h for the sparse-map model.

\section{Reproducibility}
\label{sec:s_repro}

\paragraph{Seeds}
Every model is trained with seed $42$. Every evaluation draws its query views,
frame pairs and point subsamples from fixed seeds, so all methods are compared
on identical pairs.

\paragraph{Non-determinism on GPU}
Two sources of non-determinism remain after seeding. Scatter operations in the
voxel encoder sum in a non-deterministic order. On the $60$ evaluation objects,
two runs of the object model with the same seed differ in the field by less
than $3\times10^{-4}$. This difference occasionally flips a decision of
non-maximum suppression. At $K=1024$, it changes at most $2\%$ of the keypoints
of a shape and $0.1\%$ on average. The sampling of surface points goes through
the same kind of operation and is not bit-reproducible either. We repeated the
evaluation of the object model on a second machine with the same seeds. The two
runs agree on $99.7\%$ of the frame pairs for \ours{} and on $97.4\%$ for
Harris-3D, and no table cell changes by more than $0.24$ points. On real scenes,
the fields are exported once, and the localization pipeline is deterministic
given them. Across our evaluation runs, the Harris-3D results on real scenes
agree on every frame pair. These variations are small compared with the
differences that we report.

\paragraph{Sample lists and data hashes}
The files in \texttt{manifests/} list the samples of every split and give the
SHA-256 hash of every processed file in these samples. The splits are the object
training and evaluation sets, the real-scene training, development and test
crops, the sparse object and scene test sets, the object training set of the
external baselines, and the fine-tuning and ranking sets of the corner head.
Each split is also summarized by a single hash over its sorted file hashes.

\bibliographystyle{IEEEtran}
\bibliography{biblo,c1_extra}